\documentclass[1p]{elsarticle}

\usepackage[hyperfootnotes=false]{hyperref}
\journal{Computational Statistics \& Data Analysis}

\usepackage{numcompress}

\usepackage{amsmath}
\usepackage{amsthm}
\usepackage{amsfonts}
\usepackage{bm, amssymb}
\usepackage{algpseudocode}
\usepackage{url,listings}
\usepackage{IEEEtrantools,subcaption}
 \newtheorem{thm}{Theorem} 
\DeclareMathOperator*{\argmin}{arg\,min}

\begin{document}

\begin{frontmatter}

\title{Visualizing the Effects of a Changing Distance on Data Using Continuous Embeddings}
%\tnotetext[mytitlenote]{Fully documented templates are available in the elsarticle package on \href{http://www.ctan.org/tex-archive/macros/latex/contrib/elsarticle}{CTAN}.}

%% Group authors per affiliation:
\author{Gina~Gruenhage\corref{cor1}\fnref{fn1}}
\ead{gina.gruenhage@bccn-berlin.de}
\cortext[cor1]{Corresponding author}
\address{Department of Computer Science, Technische Universit{\"a}t Berlin and BCCN Berlin,\\ Berlin, Germany}
\fntext[fn1]{Full Address: Gina Gruenhage, TU Berlin, Fakult{\"a}t IV, Elektrotechnik und Informatik, Sekr. MAR 4-2, Marchsstrasse 23, D-10587 Berlin, Tel: 0049-30-314-75729, Fax: 0049-30-314-24913 }
%\fntext[myfootnote]{?}

\author{Manfred~Opper\corref{cor2}}
\ead{manfred.opper@tu-berlin.de}
\address{Department of Computer Science, Technische Universit{\"a}t Berlin, Berlin, Germany}
%\fntext[myfootnote]{?}

\author{Simon~Barthelme\corref{cor3}}
\ead{simon.barthelme@gipsa-lab.fr}
\address{CNRS, GIPSA-lab, 11 Rue des Math{\'e}matiques, 38400 Saint-Martin-d'H{\`e}res, France}

%% or include affiliations in footnotes:
%%\author[mymainaddress,mysecondaryaddress]{Elsevier Inc}
%%\ead[url]{www.elsevier.com}

%\author[mysecondaryaddress]{Global Customer Service\corref{mycorrespondingauthor}}
%\cortext[mycorrespondingauthor]{Corresponding author}
%\ead{support@elsevier.com}

%\address[mymainaddress]{1600 John F Kennedy Boulevard, Philadelphia}
%\address[mysecondaryaddress]{360 Park Avenue South, New York}

\begin{abstract}
Most Machine Learning (ML) methods, from clustering to classification, rely on a distance function to describe relationships between datapoints. For complex datasets it is hard to avoid making some arbitrary choices when defining a distance function. To compare images, one must choose a spatial scale, for signals, a temporal scale. The right scale is hard to pin down and it is preferable when results do not depend too tightly on the exact value one picked.
Topological data analysis seeks to address this issue by focusing on the notion of neighbourhood instead of distance. It is shown that in some cases a simpler solution is available. It can be checked how strongly distance relationships depend on a hyperparameter using dimensionality reduction. A variant of dynamical multi-dimensional scaling (MDS) is formulated, which embeds datapoints as curves. The resulting algorithm is based on the Concave-Convex Procedure (CCCP) and provides a simple and efficient way of visualizing changes and invariances in distance patterns as a hyperparameter is varied. A variant to analyze the dependence on multiple hyperparameters is also presented. 
A cMDS algorithm that is straightforward to implement, use and extend is provided. To illustrate the possibilities of cMDS, cMDS is applied to several real-world data sets.
\end{abstract}


\end{frontmatter}

%\linenumbers

\newcommand{\bt}{\ensuremath{\bm{\theta}}}
\newcommand{\bmu}{\ensuremath{\bm{\mu}}}
\newcommand{\bl}{\ensuremath{\bm{\lambda}}}
\newcommand{\be}{\ensuremath{\bm{\eta}}}
\newcommand{\E}{\ensuremath{\mathbb{E}}}
\newcommand{\bys}{\ensuremath{\bm{y^{\star}}}}
\newcommand{\bS}{\ensuremath{\bm{\Sigma}}}
\newcommand{\N}{\ensuremath{\mathcal{N}}}
\newcommand{\I}{\ensuremath{\mathds{1}}}
\newcommand{\R}{\ensuremath{\mathbb{R}}}
\newcommand{\M}{\ensuremath{\mathcal{M}}}
\newcommand{\Ms}{\ensuremath{\mathcal{M}_{s}}}
\newcommand{\bx}{\mathbf{x}}
\newcommand{\bxt}[1]{\mathbf{x}^{#1}}
\newcommand{\by}{\mathbf{y}}
\newcommand{\byt}[1]{\mathbf{y}^{#1}}
\newcommand{\bxh}{\hat{\mathbf{x}}}
\newcommand{\bxht}[1]{\hat{\mathbf{x}}^{#1}}
\newcommand{\byh}{\hat{\mathbf{y}}}
\newcommand{\bX}{\mathbf{X}}
\newcommand{\bXt}[1]{\mathbf{X}^{#1}}
\newcommand{\bY}{\mathbf{Y}}
\newcommand{\bD}{\mathbf{D}}
\newcommand{\bDX}{\mathbf{\tilde{D}}}
\newcommand{\bDt}[1]{\mathbf{D}^{#1}}
\newcommand{\bDXt}[1]{\mathbf{\tilde{D}}^{#1}}

\renewcommand{\a}{\alpha}
\renewcommand{\b}{\beta}
\newcommand{\rep}{(\cdot)}
\newcommand{\simon}[1]{   {\bf \color{blue} #1} \normalcolor }
\providecommand{\norm}[1]{\lVert#1\rVert}
\section{Introduction}

\label{sec:introduction}
The notion of distance is at the core of data analysis, pattern recognition and machine learning: most methods need to know how similar two datapoints are. The choice of distance metric is often a hidden assumption in algorithms. For complex data, distance or similarity are not uniquely defined. On the contrary, they can be arbitrary to some extent \citep{Carlsson2009}. It is, for example, often possible to describe signals on different temporal or spatial scales, and distance functions will give a certain scale more weight than another. Each datapoint might describe several features, and there is often no unique, optimal way to weigh the features when computing a distance measure: are two individuals more alike if they have similar eye colour or hair colour, or do we think the shape of the nose matters most?

There are ways around that problem. One is to select the distance function that is best adapted to the task at hand, for example the one that gives the best performance in classification (this is effectively what is done in kernel hyperparameter selection \citep{Scholkopf2002}). Another is to give up on distance and rely instead on the weaker notion of neighbourhood \citep{Lum2013}. 

\begin{figure}[h]
\centering
  \includegraphics[width=3.25in]{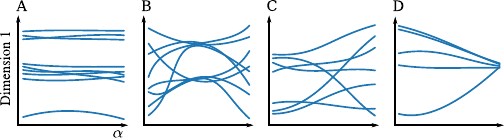}
\caption{Sketches of different effects on the data structure that emerge when varying a hyperparameter in a distance function. The $x$-axis shows the hyperparameter $\alpha$, the $y$-axis is the embedding dimension. A) Invariance: patterns hold independent of the hyperparameter. B/C) Structure emerges only for certain values of the hyperparameter. C) Declustering: clusters are lost with increasing hyperparameter. D) Information loss: Structure collapses with increasing hyperparameter.}
\label{fig:sketch}
\end{figure}

We argue here that a third option is available. One may study how the shape of the data evolves under a change in the distance metric by representing the data in lower dimension. We suppose that a family of distance functions $d_\alpha(x,y)$ is defined by varying a hyperparameter $\alpha \in [0,1]$, where $\alpha$ can represent, for example, different scales or the mixing proportion of features. Please note that $\alpha$ does not have to be defined on this interval, but it seems natural to start with a setting that is familiar from, e.g.,  convex combinations. Suppose that for a given level of $\alpha$ the relative distances between datapoints are well described by representing the datapoints as points on the line. As we vary $\alpha$ the points will move, so that each point now describes a curve. Many scenarios are possible, and we sketch them in Fig.~\ref{fig:sketch}. We may have full or partial \emph{invariance}: patterns in the data that hold regardless of the value of the hyperparameter (Fig.~\ref{fig:sketch}A). On the other hand, the structure in the data may appear only for certain values of $\alpha$ (intermediate values in Fig.~\ref{fig:sketch}B and rather small values in C), indicating that these values are more useful than others for characterizing the data. Analyzing the evolution of structures in the data might reveal interesting dependencies, for example, declustering (Fig.~\ref{fig:sketch}C) or loss of information (Fig.~\ref{fig:sketch}D).

To visualize the effects of varying the distance function we suggest to embed data into a space of smooth curves, forming what we call continuous embeddings: in continuous embeddings each datapoint is embedded as a smooth curve in  $\R^d$. We will show that this approach is quite general.

Our implementation of continuous embeddings is based on multi-dimensional scaling (MDS), one of the most widely-used tools for dimensionality reduction \citep{Buja2002,Buja2008}. MDS builds on the pairwise relation between single data points and has an intuitive way of characterizing the structure in high-dimensional data.
MDS supposes that one has distance information available, that is, we can characterize the data by a distance matrix. MDS seeks to find a set of points in a low dimensional Euclidean space, such that the Euclidean distances between points approximate the original distances. An exception is \emph{spherical} MDS, where the embedding is constrained to a spherical manifold. MDS goes back to the 1950s, when it was first introduced as classical scaling \citep{Torgerson1952}. In classical scaling, the distance matrix is transformed to a matrix of inner products from which an embedding can be computed using eigendecompositions \citep{Torgerson1952,Torgerson1958,Gower1966}. 
Classical scaling finds a perfect embedding when the data can indeed be embedded exactly, but in all realistic cases distance matrices are not exactly Euclidean and \emph{distance} scaling is more appropriate.
 \citet{Kruskal1964} introduced distance scaling by defining a cost function, \emph{Stress}, that directly measures the error between original and embedding distances. This cost function is then optimized over the space of embedding matrices which can be done using gradient descent.
Since the early work on MDS many other variants and optimization solutions have been discussed. So called non-metric variants of MDS seek to only recover the ranks of distances \citep{Shepard1962}.  \citet{Ramsay1977,Ramsay1978} introduces a statistical model for MDS, allowing for a maximum likelihood estimate. This approach is implemented in {\em Multiscale} \citep{Ramsay1978b}. Other MDS variants based on Stress include Sammon's mapping \citep{Sammon1969}, elastic stress \citep{McGee1966}, multidimensional unfolding \citep{Borg2005} and local MDS \citep{Chen2009}. Isomap \citep{Tenenbaum2000} is also related to MDS. Here, distances are computed as geodesic distances on a manifold, which are then embedded with classical scaling. In terms of optimization one of the most popular approaches is SMACOF, a majorization method for MDS \citep{Guttman1968,DeLeeuw1977,DeLeeuw1977b,DeLeeuw1988}. 

Here, we introduce a continuous version of MDS (cMDS) by adding a smoothing penalty to the MDS cost function.
Similar ideas have been used in the visualization of dynamic networks. A network is commonly represented as a graph. A 2D embedding of a static graph is often constructed using MDS or similar methods \citep{Kamada1989,Gansner2005}. In the dynamical context, where a graph is measured over time, it is important to preserve the so-called ``mental map'' when jumping from one timepoint to the next \citep{Misue1995}. Early work on such {\em controlled stability} was done by \citet{Boehringer1990,North1996}. \citet{Brandes1997} developed a more rigorous formulation of controlled stability based on regularization in a Bayesian framework. There have been three different approaches to the problem of preservation of the mental map: aggregation, anchoring and linking. 
In aggregation methods, the graph is aggregated into an average graph which is then visualized with a static layout algorithm \citep{Brandes2003,Moody2005}. Anchoring methods use auxiliary edges which connect nodes to stationary reference positions \citep{Brandes1997,Diehl2002,Frishman2008}. In linking, edges are created that connect instances of a single vertex over time. The resulting graph is then visualized using standard methods \citep{Erten2004,Erten2004b,Dwyer2006}. Linking has been formulated in more rigorous ways in terms of regularized cost functions \citep{Baur2008,Brandes2012a}. \citet{Xu2013} introduce an additional grouping penalty. \citet{Brandes2012a} provide a good overview on dynamic graph layout. Another approach to dynamic embeddings is an extension of the Hoff latent space model \citep{Sarkar2005}. In \emph{functional MDS} \citep{Masahiro2003}, individual solutions to the MDS problems are rotated in a secondary step to minimize the length of the curves. This does not allow for control of smoothness versus stress, which is true for most methods described above. 

All these approaches and contributions in the field visualize temporal developments. We show here that continuous embeddings can be applied to a great variety of data, going far beyond the visualization of temporal dynamics in graphs. In particular, the continuous variable can be used to visualize artificial dynamics. This makes the method very general and is especially useful in the analysis of families of distance functions and their effects on data structure. Continuous embeddings are thus a tool for making an informed choice of the distance metric for use in further analyses. 

We show how continuous embeddings can be efficiently computed using the Concave-Convex Procedure (CCCP) for optimization \citep{Yuille2003}. The resulting algorithm is a simple iterative procedure, in which the inner loops are nothing more than least squares regression with smoothing splines. We prove that the algorithm always leads to a stationary point. This goes further than other proofs \cite{Sriperumbudur2009,Yen2012} because the cost function is nondifferentiable at certain points and doesn't share the directional derivative with the upper bound at those points. We provide an R package for cMDS which is available on github: \href{https://github.com/ginagruenhage/cmdsr}{github.com/ginagruenhage/cmdsr}.  We illustrate the results of cMDS with several examples. We compare cMDS to a method based on $k$-means to exemplify that cMDS provides a more informative way of understanding the data structure. We show that cMDS leads to novel forms of data visualization and enhances the analysis of various meta-effects in data, such as hierarchy levels in hierarchical clustering, weighting of different distance measures and consensus requirements across subjects. Furthermore, we point out that quantitative analyses on cMDS results are possible and useful. We show that cMDS is especially well-suited to dynamic and interactive contexts \citep{Cook2007}. We provide several examples of interactive, web-based visualizations based on cMDS.

\section{Methods}

In order to present the cMDS algorithm, we introduce some notations and definitions. We describe the objective of the algorithm and present the cost function that we need to optimize. Optimization can be done in a coordinate-wise manner and we present the optimization of single coordinates via the Concave-Convex Procedure and pseudocode of the full algorithm in Section \ref{sec:findConfig}. In the subsequent section, we prove that the optimization of single coordinates is first-order optimal, i.e. it leads to stationary points of the original cost. We also show that this directly entails first order optimality of the full cost. 

We start by setting the notations. The original data or objects are denoted by $\mathbf{s}_1(\alpha) \,\ldots\, \mathbf{s}_N(\alpha)$, where $N$ is the number of objects in the data. The objects are defined in an arbitrary metric space, e.g. $\R^D$. The parameter $\alpha$ measures a continuous dimension. We will see that objects, such as images or networks, can be endowed with a continuous dimension, for example by examining them at different scales, so that scale plays the role of the continuous parameter. At this point we would like to note that we discretize all equations from the beginning, because ultimately, in the implementation, the hyperparameter has to be discretized. It would certainly be possible to develop all the mathematics in a continuous matter. We prefer to present the mathematics in such a way that the equations in the paper can be used as a direct reference for implementation. However, important equations, such as the cost function and the penalty are given in their continuous form as well, to improve the understanding of the problem definition.
Thus, $\alpha$ is represented on a grid $\alpha_1 \,\ldots\, \alpha_T$. We use $T$ to denote the maximum value of $\alpha$ since time is the most natural framework to think about the hyperparameter. 
With $f(\cdot)$ we refer to function values at all grid points while $f(k)$ refers to a function value at a specific value of $\alpha$. The objects are endowed with a distance function $d\left(x,y\right)$. The appropriate distance function depends on the data-space and on the nature of the problem. 
Given a distance measure, we can define a distance array, $\bDt{\rep} \in \R^{T\times N\times N}$, where the entry $d_{ij}^k$ holds the distance between objects $\mathbf{s}_i$ and $\mathbf{s}_j$ at $\alpha_k$. We assume that the distances between datapoints give a good summary of the patterns in the data. The goal of cMDS will be to extract these patterns.

\subsection{Objective}

The objective of cMDS is to retrieve curves or manifolds in $\R^d$, $d<<D$ which we denote as $\bxt{\rep}_1 \,\ldots\, \bxt{\rep}_N \in \R^{T\times d}$, such that the evolution of distances between the curves represents the evolution of distances between the datapoints.
When we talk about curves, we mean the technical differential geometry sense of the word, namely, a curve is a 1-dimensional manifold in $\R^d$. The curves are represented as a configuration array $\bXt{\rep} \in \R^{T\times d\times N}$ and for a given time-point $\alpha_k$ $\bXt{k}$ is a $d \times N$ matrix in which each column represents the coordinates of a curve at time $\alpha_k$. 
 In $\R^d$ we measure the distance between two curves at $\alpha_k$ with the Euclidean distance. Thus, for each configuration, we have a distance array $\bDXt{\rep} \in \R^{T\times N\times N}$ such that $\tilde{d}_{ij}^k = \norm{ \bxt{k}_i - \bxt{k}_j}_2$. To denote that $\bDXt{\rep}$ is computed from $\bXt{\rep}$, we occasionally write $\bDXt{\rep}\left(\bXt{\rep}\right)$. These are the approximate distances given by our embedding, and the objective is to make the approximate distances as close as possible to the real distances. A natural expression of that objective is the following cost function, which quantifies the distortion of the embedding:

\begin{equation}
\mathcal{L}\left(\bXt{\rep},\bDt{\rep} \right) = \int_0^1 \left(\norm{\bDXt{\a}(
\bXt{\a})-\bDt{\a}}_F\right)^2 \mbox{d}\a,
\end{equation}
where $\norm{\cdot}_F$ denotes the Frobenius norm. This is the MDS cost function for continous data. Its discretized form is:

\begin{equation}\label{eq:distortion}
\mathcal{L}\left(\bXt{\rep},\bDt{\rep} \right) = \sum_{k=1}^T \left(\norm{\bDXt{k}(\bXt{k})-\bDt{k}}_F\right)^2.
\end{equation}
 
In practice and for real datasets, MDS is a highly non-convex optimization problem with multiple local minima. Additionally, since distances are invariant to rotations, translations and symmetries, so are the MDS embeddings. Minimizing (\ref{eq:distortion}) is equivalent to solving MDS problems for different values of the hyperparameter individually. The individual problems are known as \emph{Kruskal-Shephard} scaling \citep{Kruskal1964}. This would result in solutions that are independent for different $\alpha_k$ and thus might lie in quite different local minima. However, one would expect that slight changes in the hyperparameter lead to only slight changes in the embedding. To solve this problem, we will require that each curve is continuous and smooth, which can be achieved by adding a suitable penalty function $\Omega(\bXt{\rep})$ to the cost. The effect of the smoothing penalty can be interpreted as the goal of tracing one particular local minimum across different values of the hyperparameter. This results in the cost function

\begin{equation}\label{eq:cMDS-objective}
\mathcal{C}\left(\bXt{\rep},\bDt{\rep}\right) = {\mathcal{L}\left(\bXt{\rep},\bDt{\rep}\right) +\lambda \, \Omega\left(\bXt{\rep}\right) }.
\end{equation}
Classical spline penalties \citep{Ramsay1997} are particularly convenient. We introduce the penalty in its continous form:

\begin{equation} 
  \Omega\left(\bX(\a)\right)  =   \int_0^1 \left\|\frac{\partial^2\bx(\a)}{\partial \a^2}\right\|\mbox{d}\a.
\end{equation}
 In practice, we work with discrete one-dimensional manifolds in $\R^d$ for which the penalty reads

\begin{IEEEeqnarray}{rCl} 
  \Omega\left(\bXt{\rep}\right) & = & \sum_d \text{diag}\left( \sum_i (\bxt{\rep}_i)^T\, (\mathcal{D}^{(2)})^T \mathcal{D}^{(2)}\, \bxt{\rep}_i\right) \nonumber\\
  & = & \sum_d \text{diag}\left(\sum_i (\bxt{\rep}_i)^T\, \mathbf{M} \, \bxt{\rep}_i \right),
\end{IEEEeqnarray}
where $\mathcal{D}^{(2)}$ denotes the discrete second order differential operator. 
The parameter $\lambda$ controls how strongly the roughness of the curves is penalized. For $\lambda = 0$, we recover the classical cost function of MDS, with the extension that we have $T$ separate MDS problems, one for each value of $\alpha$. For $\lambda \rightarrow \infty $ the resulting curves are straight lines, independent of the original data. The value of $\lambda$ is easy to set by visual inspection. It should simply be large enough to result in fairly smooth curves, while keeping the distortion reasonably small. A reasonable strategy for setting $\lambda$ automatically is then to maximize $\lambda$, under a constraint on the quality of the embedding. This quality can be measured in various ways \citep{Kaski2011,Mokbel2013}. The effect of the smoothing parameter $\lambda$ is shown in a supplemenatry material file.

\subsection{Optimization via the Concave-Convex Procedure (CCCP)}
\label{sec:findConfig}

The cMDS algorithm optimizes the cost in a curve-by-curve manner. This is possible since the cost is a sum over costs per curve. Thus, we have an outer loop over curves and an inner loop that performs conditional optimization on $\bxt{\rep}_i$. That is, we assume that all curves $\bxt{\rep}_j, j\ne i$ are fixed. The inner loop is a Maximization Minimization procedure. Specifically, we use the Concave-Convex Procedure (CCCP) \citep{Yuille2003}. This way, each minimization step is a simple spline regression. In effect, the algorithm only needs to compute spline regressions with surrogate data.  

We now outline the usage of CCCP for the optimization of a single curve. We expand the first term of the cost function to identify the convex and concave parts. We denote the cost of a single curve as $f(\cdot)$: 

\begin{IEEEeqnarray}{rCl}
  f\bigl(\bxt{\rep}_i\bigr) & = & \sum_{kj} \left( \left\| \bxt{k}_i - \bxt{k}_j\right\|_2 - d_{ij}^k \right)^2+ \lambda (\bxt{\rep}_i)^T\, \mathbf{M}\, \bxt{\rep}_i \nonumber\\
 & = & \sum_{kj}(\bxt{k}_i - \bxt{k}_j)^T(\bxt{k}_i-\bxt{k}_j) \nonumber \\
 &&- \sum_{kj} 2\,d_{ij}^k \left\|\bxt{k}_i - \bxt{k}_j\right\| + \sum_{kj} (d_{ij}^k)^2 \nonumber\\
&& +\> \lambda\, \bigl(\bxt{\rep}_i\bigr)^T\, \mathbf{M}\, \bxt{\rep}_i \nonumber\\
& = & f_{vex}\bigl(\bxt{\rep}_i\bigr) + f_{cave}\bigl(\bxt{\rep}_i\bigr) + \text{const},
\end{IEEEeqnarray}
where 

\begin{align}
  &f_{vex}\bigl(\bxt{\rep}_i\bigr) = \sum_{kj} (\bxt{k}_i- \bxt{k}_j)^T(\bxt{k}_i-\bxt{k}_j) + \lambda\, (\bxt{\rep}_i)^T\, \mathbf{M}\, \bxt{\rep}_i\\
\intertext{and}
  &f_{cave}\bigl(\bxt{\rep}_i\bigr) = - \sum_{kj} 2\,d_{ij}^k \left\|\bxt{k}_i - \bxt{k}_j\right\|.
\end{align}
In the optimization, we can omit the constant term that doesn't depend on $\bxt{\rep}_i$.
The iterative CCCP algorithm $(\bxt{\rep}_i)^{t-1} \mapsto (\bxt{\rep}_i)^{t}$ is given by

\begin{equation}
  (\bxt{\rep}_i)^{t} = \argmin_{\bxt{\rep}_i}\;\; u \bigl(\bxt{\rep}_i, (\bxt{\rep}_i)^{t-1}\bigr),
\label{eq:cccp-def}
\end{equation}
where the convex upper bound is computed by taking the first-order Taylor expansion of the concave part of $f$.
The concave part is, however, non-differentiable at $\bxt{k}_i = \bxt{k}_j$. We thus work with a modified subdifferential.

\begin{IEEEeqnarray}{rCl}
  \IEEEeqnarraymulticol{3}{l}{\vec\nabla^{SD} f_{cave}\left((\bxt{k}_i)^{t-1}\right)}\nonumber  \\ 
 \quad & = & 2 \sum_k\sum_i d_{ij}^k \begin{cases} \frac{(\bxt{k}_i)^{t-1} - \bxt{k}_j}{\left\|(\bxt{k}_i)^{t-1} - \bxt{k}_j\right\|} & \text{if $ (\bxt{k}_i)^{t-1} \ne \bxt{k}_j$} \\
\left\{\mathbf{u} :  \left\| \mathbf{u} \right\| = 1 \right\} & \text{if $(\bxt{k}_i)^{t-1} = \bxt{k}_j$} \end{cases}.
\label{eq:Subdiv}
\end{IEEEeqnarray}
In the latter case, $\mathbf{u}$ is chosen randomly. The usual definition of the subdifferential \cite{Rockafellar1970} would yield a less strong condition on $\mathbf{u}$, namely $\left\| \mathbf{u} \right\| \le 1$. We choose the stronger variant to achieve maximum descent in each optimization step.
We introduce surrogate points $\bxht{\rep}_j$.

\begin{equation}
  \bxht{k}_j = \bxt{k}_j + d_{ij}^k \begin{cases} \frac{(\bxt{k}_i)^{t-1} - \bxt{k}_j}{\left\|(\bxt{k}_i)^{t-1} - \bxt{k}_j\right\|} & \text{if $ (\bxt{k}_i)^{t-1} \ne \bxt{k}_j$} \\
\left\{\mathbf{u} :  \norm{\mathbf{u}} = 1 \right\} & \text{if $(\bxt{k}_i)^{t-1} = \bxt{k}_j$} \end{cases},
\end{equation} 
where, again, $\mathbf{u}$ is chosen randomly.

The resulting upper bound for the original cost is 

\begin{IEEEeqnarray}{rCl}
  u\bigl( (\bxt{\rep}_i)^{t}, (\bxt{\rep}_i)^{t-1}\bigr) & =& \sum_{kj} \bigl((\bxt{k}_i)^t - \bxt{k}_j\bigr)^T\bigl((\bxt{k}_i)^t-\bxt{k}_j\bigr) \nonumber\\
 && -\> 2 \sum_{kj} d_{ij} \norm{(\bxt{k}_i)^{t-1} -\bxt{k}_j} \nonumber\\
 &&  -\> 2 \sum_{kj} \bigl((\bxt{k}_i)^t - (\bxt{k}_i)^{t-1}\bigr)\bigl(\bxht{k}_j - \bxt{k}_j\bigr) \nonumber\\ &&+\> \lambda\, \bigl((\bxt{\rep}_i)^t\bigr)^T\, \mathbf{M}\, (\bxt{\rep}_i)^t.
\label{eq:upperbound}
\end{IEEEeqnarray}

The updating step is thus a simple spline regression and can be performed analytically. Spline regression involves a matrix inversion with cost $\mathcal{O}(T^3)$, but the inverse needs to be computed only once. Once the inverse has been obtained the cost of the spline regression becomes $\mathcal{O}(T^2)$, with further savings possible using sparse matrix techniques. For simplicity of notation, we rewrite the embedding points $\bxt{\rep}$ as column vectors in $\R^{T\cdot d}$, $\vec\bx_i = \mathbf{vec} \, \bxt{\rep}_i$.

\begin{equation}
  \label{eq:update}
  \vec\bx_i^{t} = \left(N\cdot \mathbf{E}_{T \cdot d} + \lambda \mathbf{E}_d \otimes \mathbf{M}\right)^{-1}\left(\sum_j \vec\bxh_j\right),
\end{equation}
where $\mathbf{E}_d$ is the identity matrix in $\R^{d\times d}$.

\citet{Agarwal2010} introduce the same surrogate points based on geometrical considerations for metric (non-continuous) MDS. There, they are defined as projecting $\bx_i$ on the sphere $\mathcal{S}_j$ with center $\bx_j$ and radius $d_{ij}$. In our formulation, they arise naturally from a natural splitting of the cost function into convex and concave parts. Additionally, using the subgradient circumvents the problem of the surrogate points being undefined for $\bx_i = \bx_j$, which \citet{Agarwal2010} do not address.

Together with the iteration over the curves $\bxt{\rep}_i$ we have everything we need for the cMDS algorithm (see Fig.~\ref{alg:cmds}). The complexity of the algorithm is $\mathcal{O}(dT^2
 N^2)$ per iteration of the outer loop, since its function \textsc{MM} has complexity $\mathcal{O}(dT^2N)$ due to the matrix product in the spline regressions.

\begin{figure}
\begin{algorithmic}[0]
\Function{cMDS}{$\bDt{\rep},\bX^{\rep}_0,\lambda,\mathbf{M},\delta$}
\State maxIter = 50
\State k = 0
   \While{$k < \text{maxIter}$}
   %\State $\epsilon \gets \bXt{\rep}$
      \For{($i = 1:N$)} 
      \State $\bxt{\rep}_i \gets$ \Call{MM}{$\bDt{\rep},\bXt{\rep},i,\text{params}$}
      \EndFor
      \State $k \gets k+1$
   \EndWhile
   \State \textbf{return} $\bXt{\rep}$
\EndFunction

\vspace{0.5cm}
\Function{MM}{$\bDt{\rep},\bXt{\rep},i$, params}
   \Repeat  
      %\State $\epsilon \gets \bxt{\rep}_i$
      \State $\bxht{\rep} \gets $ \Call{Majorize}{$(\bXt{\rep})^{t-1},\bDt{\rep},i,\text{params}$}
      \State $(\bXt{\rep})^t \gets$ \Call{Minimize}{$(\bXt{\rep})^{t-1},\bxht{\rep},i,\text{params}$}
      \Until{$\left(\frac{\frac{1}{N} \sum_i\left|(\bxt{\rep})^{t-1}- (\bxt{\rep}_i)
^t\right|}{\frac{1}{N}\sum_i\left|(\bxt{\rep}_i)^t\right|} > \delta\right)$} 
      \State \textbf{return} $\bxt{\rep}_i$   
\EndFunction
\end{algorithmic}
\caption{The cMDS algorithm. The maximum number of iterations is set to 50. In practice, this is a sufficient number of iterations for the outer loop. The parameters passed to the MM function include, for example, $N$, $d$,$T$,$\lambda$, weights (if used), the penalty matrix $\mathbf{M}$ and the error tolerance.}
\label{alg:cmds}
\end{figure}

The resulting algorithm can also be interpreted as a majorization-minimization algorithm \citep{DeLeeuw1977,DeLeeuw1994,Hunter2004} which is true for any CCCP algorithm \citep{Sriperumbudur2009}. The majorization entails the computation of the surrogate points, while minimization is the computation of the updating step. 

\subsection{First-order optimality of the cMDS algorithm}
\label{sec:proof}

The cMDS algorithm has the structure of a coordinate-descent method, with an outer loop that improves the configuration curve-by-curve, and an inner loop that optimises the objective for a given curve (using CCCP). As we will see it is difficult to make very strong statements about the convergence of a coordinate-descent algorithm on a non-convex, non-differentiable objective such as ours. The guarantee we can offer is that, if cMDS converges, then it converges to a local minimum or saddle point. We begin with a study of the inner loop, and discuss the outer loop in the next subsection. 

\subsubsection{First-order optimality of the inner loop}
Since CCCP has become an important tool in machine learning, there is a vast literature on convergence proofs for CCCP in general. \citet{Sriperumbudur2009} analyze the convergence of CCCP, for smooth and differentiable cost functions. In this case, the authors prove global convergence, in the sense that the algorithm arrives at a stationary point, i.e. a local minimum or saddle point, from any initialization point. The cMDS cost function, however, is nonsmooth and nondifferentiable at $\bxt{k}_i = \bxt{k}_j$. 
\citet{Yen2012} focus on convergence rate of CCCP and work with nonsmooth objective functions. They connect CCCP to more general block coordinate descent methods. They use results on convergence of coordinate descent on nonconvex and nonsmooth problems \citep{Tseng2009} by showing that CCCP is an instance of block coordinate descent. However, the authors constrain the class of objective functions to which their proof applies. The nonsmooth part should be convex piecewise linear, which is not the case for cMDS.
\citet{Razaviyayn2013} extend block coordinate descent methods to inexact block coordinate descend. This allows the authors to treat nondifferentiable and nonconvex objective functions. The authors unify convergence results for various methods such as CCCP and Expectation Maximization. We follow ideas of this paper to show that a limit point of the optimization chain of a single curve in cMDS via CCCP is a stationary, first-order optimal point of the cost. 

For readability we denote $x = \bxt{\rep}_i$ in the following.
It is easy to see that, by optimizing single curves via minimization of an upper bound, we monotonically decrease the original cost. 

\begin{equation}
  \label{eq:decr.seq}
  f\left(x^{1}\right) \geq f\left(x^{2}\right) \geq f\left(x^{3}\right) \geq \dots\,.
\end{equation}
However, from this it is not yet clear (assuming convergence of the algorithm) that the resulting configuration is a stationary point of the original cost, since the algorithm could produce a non-increasing sequence that does not tend to an optimal point.  Therefore, we need to show that the limit point $z$ of the inner loop is first-order optimal, i.e.
$\vec{\nabla} f(x) \bigl|_{x=z} = 0$. Then, we prove that this entails first-order optimality of the inner loop. Strictly speaking, this only holds with probability one, with respect to the randomisation step
in (\ref{eq:Subdiv}), which chooses the direction of the next step when an iterate ends on one of the other data points.
We will first show that with probability one,  such a point will never be a fixed point of the algorithm.
 For simplicity, we consider this for the standard MDS cost function without the penalty term. 
The updating step is then

\begin{equation}
  \bx_i^t = \frac{1}{N} \sum_j \bxh_j.
\end{equation}
Now we assume, without loss of generality, that the current iterate $\bx_i^{t-1} = \bx_N$. Using the definition of the auxiliary points, we have:

\begin{equation}
  \bx_i^t = \frac{1}{N}\left( \sum_j \bx_j + \sum_{j=1}^{N-1} d_{ij}\frac{\bx_N-\bx_j}{\left\| \bx_N-\bx_j \right\|} + d_{iN} \,\mathbf{u}\right).
\label{eq:lin_t}
\end{equation}
If this point was a fixed point of the algorithm, i.e. $\bx_i^{t} = \bx_i^{t-1}$, 
there would be only one single direction $\mathbf{u}$ which fulfills the resulting linear equation (\ref{eq:lin_t}). But this event has probability zero when $\mathbf{u}$ is chosen randomly.

We thus know that for any fixed point, with probability one, $u(\cdot,z)$ and $f(\cdot)$ are differentiable at $x=z$. Hence, with probability one, for any fixed point $z$ of the algorithm the following condition on the derivative holds: 
%\footnote{The directional derivative of a function $f \,:\,\mathcal{D} \rightarrow \R$, where $\mathcal{D}$ is a convex set, is defined as \[f'(x;d) = \liminf_{\lambda 0} \frac{f(x+\lambda d) - f(x)}{\lambda}\].} 

\begin{equation}
\vec{\nabla} u(x, z)\biggl|_{x = z} = \vec{\nabla} f\left(z\right).
\label{eq:con_deriv}
\end{equation} 

\begin{thm}
\label{thm:innerloop}
Every limit point of the iterates generated by (\ref{eq:update}) is first-order optimal.
\end{thm}
\begin{proof}
  Let us assume that a subsequence ${x^{t_j}}$ exists which converges to a limit point $z$. 

  \begin{equation*}
    u(x^{t_{j+1}}, x^{t_{j+1}}) = f\left(x^{t_{j+1}}\right) \leq u(x^{t_{j+1}}, x^{t_{j}}) \leq u(x, x^{t_{j}}) \; \forall x.
  \end{equation*}
 The equality follows from the fact that $u(y, y) = f\left(y\right)$. The first bound follows from $u(x,y) \geq f(x)$ and the last inequality is due to the optimality of $x^{t_{j+1}}$.
We now perform the limit $j \rightarrow \infty$ and arrive at

\begin{equation*}
u(z,z) \leq u(x,z) \; \forall x .
\end{equation*}
Thus $u(\cdot,z)$ has a local minimum at $x=z$. This implies that

\begin{equation*}
  \vec{\nabla} u(x,z)\biggl|_{x = z} = 0,
\end{equation*}
since $u(\cdot,z)$ is a convex differentiable function.
Combining this with (\ref{eq:con_deriv}) we obtain

\begin{equation*}
  \vec{\nabla}f(z) = 0.
\end{equation*}
\end{proof}

\subsubsection{Convergence of the outer loop}

The outer loop has the structure of a (block)-coordinate descent algorithm (see \citep{Wright2015}, for a review). Surprisingly, there are few results in the literature  that could be used to guarantee convergence of coordinate-descent on a non-differentiable, non-convex objective function. There are some examples to the contrary: \citet{Powell1973} gives an example of a non-convex differentiable objective that leads coordinate descent to loop forever between suboptimal points. 

In our particular case, however, we have established that points of non-differentiability cannot be fixed points of the inner loop, and that the set of fixed points of the inner loop are first-order optimal (meaning the gradient of the cost with respect to curve $x_i$ is 0). A global fixed point would therefore imply that all conditional gradients are null, meaning that the fixed point is first-order optimal. Further, since the algorithm produces a strictly non-increasing sequence of cost values, it can only converge to a local minimum or saddle-point (if it converges at all). 

This proof does not exclude the possibility of a limit cycle, meaning that there could be several limit points with the same cost which the algorithm visits in turns. However, in our implementation, we determine convergence based on the stationarity of the configuration and not of the cost. Thus, if the inner loop converges, it converges to a unique configuration which is a stationary point. If the algorithm did indeed jump between several limit points, it would not converge and stop after a certain number of iterations. However, this has never happened in our experience.

% \begin{thm}
%   The cMDS algorithm has first-order optimal limit points.
% \end{thm}
% \begin{proof}
%   The cost function is symmetric, meaning that the cost of points $i$ and $j$ is the same as the cost of points $j$ and $i$. It is easy to see that in this case, coordinate-wise first-order optimality yields overall first-order optimality. With Lemma \ref{lem:innerloop} we know that the inner loop of the algorithm yields such coordinate-wise optimality. Thus, we get first-order optimality of the limit points generated by the cMDS algorithm.
% \end{proof}
 
Beyond the fact that the algorithm does converge to an appropriate point, it would be interesting to prove some results on the speed at which it converges. Unfortunately, local convergence is very difficult to study in our case since most techniques rely on the assumption that the cost function becomes quadratic in a neighbourhood of an optimum. Due to the invariances inherent in the MDS cost function, it is not clear at all that a local quadratic model is in any way appropriate.

\subsection{cMDS with more than one hyperparameter}
In certain cases it might be interesting to look at the effects of varying more than one hyperparameter. An extension of cMDS is thus the use of two or more hyperparameters, $\a$ and $\b$. This allows to consider, for example, time plus an additional hyperparameter such as scaling or weighting.  Having multiple hyperparameters effectively only changes the penalty matrix used in the spline regressions: instead of a penalty matrix appropriate for parametric curves in $\mathbb{R}\rightarrow\mathbb{R}^d$, we need a penalty appropriate for curves in $\mathbb{R}^m\rightarrow\mathbb{R}^d$, where $m$ is the number of parameters.

As an example, take the case where we have two hyperparameters, $m = 2$. Suppose we have a grid $\a_1 \,\ldots\, \a_{T_\a} \times \b_1 \,\ldots\, \b_{T_\b}$. Then, our configuration matrix is $\bXt{\rep\rep} \in \R^{T_\a \cdot T_\b \times d \times N}$. If $T_\a \neq T_\b$, we have two matrices for the discrete second order differential operator, $\mathbf{M}_\a$ and $\mathbf{M}_\b$, of appropriate dimensionality. Thus, we get the following penalty matrix for two hyperparameters:

\begin{equation}
  \label{eq:two_hyperparameters}
  \mathbf{M} = \mathbf{M}_\b \otimes \mathbf{E}_{T_\a} + \mathbf{E}_{T_\b}\otimes \mathbf{M}_\a,
\end{equation}
defining a separable penalty across the two hyperparameters. For more on penalty matrices, see for example \citep{Ramsay1997}.

The results can be visualised by selecting certain slices of the grid. In the case of 1-dimensional embeddings, selecting a certain value for one hyperparameter leads to the visualization of curves that depend on the other hyperparameter. In case of 2-dimensional embeddings, one can select a certain value of one hyperparameter and then analyze the corresponding, possibly animated, scatterplot that varies with the second one.

\subsection{Initialization}
\label{sec:init}
The cMDS algorithm needs to be seeded with a starting configuration. The optimization works with a random initialization, but performance can be improved with a more structured approach. 

One possibility is to perform classical scaling \citep{Kruskal1964,Torgerson1952} or other standard MDS methods such as SMACOF \citep{DeLeeuw1977,DeLeeuw2009} for each value of $\alpha$. However, the separate solutions might be difficult for cMDS to "glue" together and thus form a poor initialization. A more robust variant is initialization with an aggregated solution. To obtain this solution we average each curve $\bxt{\rep}_i$ over all values of $\alpha$ and then perform classical scaling on $\bar{\mathbf{X}}$. If some patterns are very strong for only a certain range of $\a$, they might influence the entire embedding via the aggregated initialization. Thus, one should consider which kind of initialization is suitable for the data at hand.

\subsection{Embedding dimension}
We would like to shortly comment on the choice of the embedding dimension. Since we aim for visualization, one only needs to choose between $d=1$ or $d=2$. One way to choose is to look at the so-called Shepard plots. In such a plot, embedding distances are plotted against original distances. For an ideal embedding, all the points in such a plot would fall on the diagonal. Thus, the spread from the diagonal can be used to judge whether the embedding is meaningful or not. If the Shepard plot looks reasonable for $d=1$, one should choose this dimension. If there is a significant improvement for $d=2$, one should choose the higher dimension.

\subsection{Variants of cMDS}
There exists multiple variants of the standard MDS problem. All of them can be implemented in cMDS. Most variants are defined over weights $\mathbf{W}^{\rep}$ in the cost function:

\begin{equation*}
  \label{eq:cost}
  \begin{split}
 \mathcal{C}\left(\bXt{\rep},\bDt{\rep}\right)
 =& \sum_{k=1}^T \norm{\mathbf{W}^k \circ \left( \bDXt{k} - \bDt{k} \right)^2}_F \\
 &+ \lambda \sum_i (\bxt{\rep}_i)^T\, (\mathcal{D}^{(2)})^T \mathcal{D}^{(2)}\, \bxt{\rep}_i.
 \end{split}
\end{equation*}
{\em Sammon's mapping} \citep{Sammon1969} can be implemented by setting $w_{ij}^k = (d_{ij}^k)^{-1}$. In {\em elastic stress}, on the other hand, we have weights $w_{ij}^k = (d_{ij}^k)^{-2}$ \citep{McGee1966}. This is equivalent to a Kamada-Kawai layout in graph visualization \citep{Kamada1989}. {\em Multidimensional unfolding} is useful when the data separates into groups \citep{Borg2005}. Then, the weights corresponding to between-group distances are set to zero. Local MDS (LMDS) \citep{Chen2009} is a slightly more complex variant. Here, the weights are set depending on local neighborhoods: If object $\mathbf{s}_j^k$ is a k-nearest neighbor of $\mathbf{s}_i^k$ (which we denote as $j \in \mathcal{N}_i$), then the corresponding term in the cost gets weight $w_{ij}=1$ and the original distance $d_{ij}$ is used. In all other cases the weight $w$ is set to a small value (not dependent on $i,j$) and $d_{ij}=D_{\infty}$, where $D_{\infty}$ is a large constant. This leads to a focus on local neighborhood structure and adds a repulsive force to avoid a 'crumbling together' of the embedding. The corresponding cMDS cost function is

\begin{equation*}
  \label{eq:cost}
  \begin{split}
 \mathcal{C}\left(\bXt{\rep},\bDt{\rep}\right)
 =& \sum_{k=1}^T \sum_{i=1}^N\sum_{j \in \mathcal{N}_i} \left( \norm{\bxt{\rep}_i-\bxt{\rep}_j}_2 - d_{ij}^k \right)^2 \\
 &+ \sum_{k=1}^T \sum_{i=1}^N\sum_{j \not\in \mathcal{N}_i} w^k \left( \norm{\bxt{\rep}_i - \bxt{\rep}_j}_2 - D_{\infty}^k \right)^2 \\
 &+ \lambda \sum_i (\bxt{\rep}_i)^T\, (\mathcal{D}^{(2)})^T \mathcal{D}^{(2)}\, \bxt{\rep}_i.
 \end{split}
\end{equation*}

LMDS introduces an additional hyperparameter via $w$, that needs to be optimized for the data at hand. 

It is also possible to use ISOMAP \citep{Tenenbaum2000} to visualize changes in manifold structure that are induced by a change in metric. To this end, one constructs geodesic distances based on different distances $d_{\alpha}\left(x,y\right)$ and subsequently applies cMDS.

\subsection{R package {\bf cmdsr}}
To make cMDS accessible we provide an R package, {\bf cmdsr}, on github \href{https://github.com/ginagruenhage/cmdsr}{(here)}. The heart of the package is the function {\bf cmds}. Functions for plotting and summarizing results are also available. To show the usability of cMDS we display exemplary function calls here:

\begin{quote}
  ComputeCmds(kDistances, kDim = 2, kLambda = 5, kWeights = ``sammon'')
%\end{quotation}
%\begin{quotation}

  ComputeCmds(kDistances, kDim = 1, kLambda = 3, kInit = ``smacof'')
\end{quote}
%\end{lstlisting}
where DL is a list of distance matrices of length $T$, k is the embedding dimension and l is $\lambda$, the regularization parameter. $W$ can be set
to use different variants of MDS, such as {\em Sammon's mapping}, {\em Kamada-Kawai} and {\em unfolding}. Different initializations are also available. There is also a built-in plotting function. For example, the Shepard plot can be plotted like this:

\begin{quotation}
  PlotCmds(ComputeCmds(kDistances), kShepard = TRUE)
\end{quotation}
Further examples can be found in the package documentation.

\subsection{Distance families}
We would like to discuss some general properties of distance families.

\paragraph{Weighted distances}
A very common case is that data can be described using different features or sets of features. In that case, we can build two distance matrices $\mathbf{D}_1$ and $\mathbf{D}_2$ on one feature respectively. Then, we can construct a weighted metric using a convex combination of the two. Using a convex combination ensures that each resulting matrix is again a distance matrix.

\begin{equation}
  \bDt{\rep} = \sqrt{ \alpha \mathbf{D}_1^2 + (1-\alpha) \mathbf{D}_2^2}\,.
\end{equation}
We present examples of weighted distance in Section \ref{sec:examples}.

\paragraph{Changing inherent dimensionality}
An interesting issue is a change in intrinsic dimensionality in the distance function. In classical scaling, low dimensional embeddings are projections of higher dimensional embeddings. For example, a 2d embedding is the projection of the 3d embedding. In reverse, this means that embedding high dimensional data in lower dimensions leads to larger distortion. In continuous embeddings, the dimensions are not stacked, as in the classical case. However, it is still true that a larger difference in dimensionality between original and embedded data leads to larger distortion. We illustrate this by mixing a distance matrix based on low dimensional data ($d=2$) with one based on high dimensional data ($d=12$). We embed this data in $d=2$. Plotting the distortion for each value of $\alpha$, as defined in (\ref{eq:distortion}), shows that, as expected, embedding the 2d data in $d=2$ yields zero distortion. The distortion increases as the high dimensional data gets more weight in the mixture. 

\begin{figure}[h]
\centering
  \includegraphics{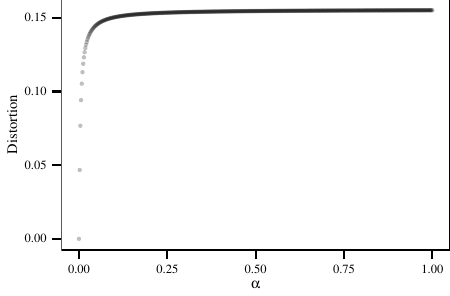}
\caption{Change of inherent dimensionality: $\alpha=0$ represents a distance matrix based on two dimensional data, $\alpha=1$ is based on 12-dimensional data. The mixture was embedded in $d=2$. It is clearly visible that, as the influence of the high-dimensional data increases, the distortion also increases. }
\label{fig:inherent_dimension}
\end{figure}

\section{Examples}
\label{sec:examples}

In the following examples we show how cMDS can be used to analyze the effect of hyperparameters of distance functions on the data. The first example is a toy example, where we illustrate how cMDS can visualize cluster structure in data and how it compares to a more basic method based on $k$-means. In the second example, we use a weighted metric: distances evolve according to the relative weight given to some of the dimensions. In the third example, we work with brain connectivity data. We vary the distance function between networks according to different thresholding rules. In the fourth example, we derive a multiscale representation of data using hierarchical clustering and visualize changes in distance over scale. 

\subsection{Comparison of cMDS to clustering}
We compare cMDS to a rudimentary method of tracking the influence of changing distances on the data structure.

An approach based on traditional methods is to perform clustering on the original data for multiple instances of the distance family $d_{\alpha}(x,y)$. To track whether clusters in the data emerge independently of the distance measure, we use the set of cluster indices $c_i$ of a specific level of $\alpha$ for all levels of $\alpha$ and compute the corresponding quality of the clustering result, for example, by comparing within and between cluster variance. By good clustering quality, we mean data that has well seperated clusters. Low quality on the other hand refers to strongly overlapping clusters. If the quality breaks down, this should be visible in the cMDS result on first glance. 

\begin{figure*}
  \begin{subfigure}{0.48\textwidth}
    \includegraphics[width = \textwidth]{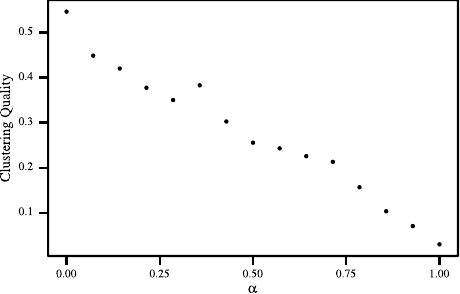}
    \caption{}
    \label{fig:clustering_kmeans}
  \end{subfigure}
  \begin{subfigure}{0.48\textwidth}
    \includegraphics[width = \textwidth]{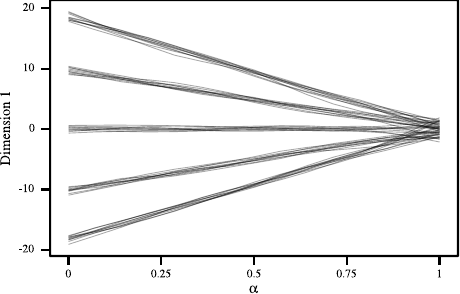}
    \caption{}
    \label{fig:clustering_cmds}
  \end{subfigure}
  \caption{(a) Clustering quality based on $k$-means clustering. We performed clustering at $\alpha=0$ and used the resulting indices at all levels of $\alpha$. One can see that the quality decreases, thus indicating that the cluster structure that is present in the beginning is not invariant. However, one cannot know what happens with the data exactly. (b) cMDS on the toy example. $\alpha=0$ represents the distance matrix based on data in five clusters in $\R^5$. $\alpha=1$ represents a random distance matrix. The results provide at a glance that clusters dissolve at larger values of $\alpha$. This information is difficult to obtain based on $k$-means clustering alone.}
\end{figure*}

We demonstrate this with an artificial example. Suppose we draw five random cluster centers in $\R^5$ and then let those cluster centers linearly collapse to zero, such that for $\alpha=0$ we have well separated cluster centers and at $\alpha=1$ all cluster centers are at the origin. In our example, we draw the cluster centers uniformly from a 5-dimensional hypercube with coordinate-wise limits of $[-10,10]$ and sample 15 points from a linear approximation between the centers and the origin. We then simulate Gaussian data at all values of $\alpha$, using the respective cluster centers and simulating 10 points per cluster. We use a variance of 0.3 to ensure that the clusters are well separated in the beginning. We then compute the corresponding distance matrices for each level of $\alpha$. Now, let us assume we don't know anything about the data. We compute a $k$-means clustering for $\alpha=0$. We use the resulting cluster indices for all levels of $\alpha$ and compute cluster quality. We use the following quality measure:

\[q = \left( \sum_i \frac{d_{max}(x_{c_i})}{d_{min}(c_i)}\right)^{-1} \]
i.e. for each cluster, we compute the maximum distance between two points in this cluster and the minimum distance from this cluster's center to another cluster center. The inverse of the sum of those values gives our quality measure. Low quality (strong overlap) of the clustering leads to low values of $q$.
We run $k$-means and compute $q$ ten times to get an average measure for cluster quality. The results are shown in Fig.~\ref{fig:clustering_kmeans}. We see that the quality of the clustering decreases with $\alpha$. From this we can conclude only that the clustering in the beginning is not present at the end. What we do not know is what the data actually looks like. Are there no clusters in the end or maybe different clusters than in the beginning? Let us now look at the cMDS embedding of the distance matrices (Fig.~\ref{fig:clustering_cmds}). Here we see that the clustering structure slowly degenerates with increasing $\alpha$ and that no other clusters arise for large $\alpha$. Thus, we gain a lot more insight about the data without having to run $k$-means many times and computing average cluster quality measures. 

\subsection{Economic and demographic descriptors of EU countries}
\label{sec:EU}

\begin{figure*}[tbph]
\centering
  \includegraphics[width=\textwidth]{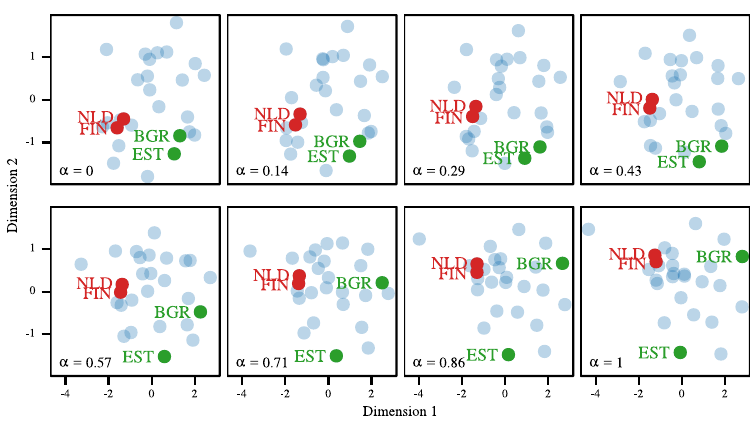}
\caption{Effects of changes in a weighted metric. Economic and demographic distance measures are weighted according to $\bD(\alpha) = \sqrt{\alpha \cdot \bD_E^2 + (1-\alpha) \cdot \bD_D^2}$. Thus, the first panel represents an embedding that is solely based on the demographic metric, while the last one is based on the economic one. In between are samples of different weights, denoted by $\alpha$. The red labels depict the neighborhood relation between the Netherlands (NLD) and Finland (FIN) which is invariant with respect to $\alpha$, while the green ones, Bulgaria (BGR) and Estland (EST), are an example of a strong dependency on $\alpha$.}
\label{fig:EUcountries}
\end{figure*}

For this example we use economic and demographic descriptors of the 27 EU countries. The data are publicly available from the {\bf gapminder} website. As economic variables we use income per capita (2008), CO$\mbox{}_2$ emissions per capita (2008) and number of granted patents per capita (2002). Demographic variables are total fertility rate, life expectancy at birth and the fraction of urban population. We scale all variables logarithmically. We then build two distance matrices, one solely based on economic variables, $\bD_E$, the other on demographic variables, $\bD_D$. Now, we put different weights on both variable groups and have a continuous range with one extreme only considering demographic variables and the other extreme only considering economic ones:

\[\bD(\alpha) = \sqrt{\alpha \cdot \bD_E^2 + (1-\alpha) \cdot \bD_D^2}\]
where $\alpha$ is between 0 and 1. We sample this continuum at $N$ different values of $\alpha$ and thus get a distance array in $R^{N \times 27 \times 27}$. 
For this example, a 1D embedding is not sufficient to capture relevant trends in the data. Thus, we give snapshots of the 2D results in Fig.~\ref{fig:EUcountries}. In this case, an interactive presentation of the results is much easier to read and thus an advantageous choice. An interactive visualization is viewable at \href{http://tinyurl.com/cMDS-demo}{http://tinyurl.com/cMDS-demo}. We also implemented an interactive web application with the R package {\bf shiny}, which is online at \href{http://ginagruenhage.shinyapps.io/EU-App}{http://ginagruenhage.shinyapps.io/EU-App} .
The 2D embedding shows that the changes in weighting have significantly different effects on the individual neighborhood relations. For example, some demographically very similar countries start diverging when economic variables are taken into account and end up far apart under the economic distance metric (e.g. Bulgaria and Estonia). Other countries stay similar, independent of the weighting of different distances (e.g. Finland and the Netherlands). These patterns are lost when deciding a distance measure a priori. Using cMDS to visualize the effects of the hyperparameter makes them easier to discover and understand. In the web application, one can also toggle quantitative analyses of the cMDS output, such as a vector graphic showing local stability of various countries. It is also possible to color countries according to their respective penalty, to judge overall stability.

\subsection{Diffusion Tensor Imaging}
\label{sec:DTI}

\begin{figure*}[tbph]
\begin{center}
%\centering
  \includegraphics[width=\textwidth]{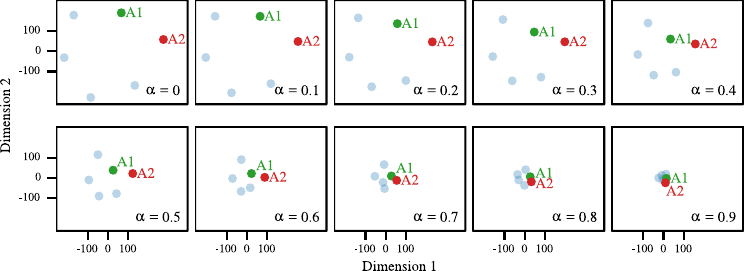}
\caption{Embedding of thresholded networks acquired by DTI and tractography \citep{Hagmann2008}. The hyperparameter $\alpha$ corresponds to the quantile of weakest connections that is ignored for inter-subject comparisons. Measurements are compared based on the resulting binary connectivity matrix using the Hamming distance between graphs. The analysis shows that the two measurements for subject A, A1 and A2, differ about as much as other pairwise comparisons. This is interesting because it highlights the relatively low reliability of DTI tractography.}
\label{fig:dti_network}
\end{center}
\end{figure*}

\begin{figure*}
\centering
  \includegraphics[width=\textwidth]{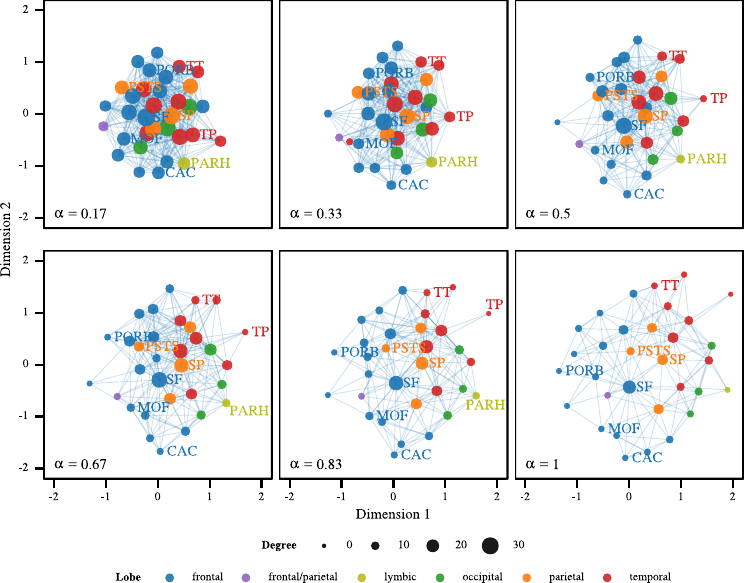}
\caption{Embedding of regional networks acquired by DTI and tractography \citep{Hagmann2008}. The different panels correspond to different threshold levels $\alpha$ when building consensus networks. The superior frontal cortex (SF) and superior parietal cortex (SP) are stable regions in the core of the network, while the parahippocampal cortex (PARH), the caudal anterior cingulate (CAC), and the transverse temporal cortex (TT) are stable peripheral regions. Some regions are less stable, e.g. the pars orbitalis (PORB), the temporal pole (TP), the medial orbitofrontal cortex (MOF) and the postcentral gyrus (PSTS).}
\label{fig:dti_concensus}
\end{figure*}

Brain regions are linked by white matter tracts, forming a network called the Connectome \citep{Koetter2005}. Diffusion Tensor Imaging (DTI) is a form of magnetic resonance imaging that can be used to find connections between brain regions (using tractography, \citep{Hagmann2003}). Here we use  data obtained by \citet{Hagmann2008}, available \href{http://www.cmtk.org/datasets/homo_sapiens_01.cff}{here}. DTI produces noisy results and it is difficult to compare individual subjects. Consequently, connectivity must often be averaged over individual subjects. There is by necessity some arbitrariness in the averaging and we show here how cMDS can be used to visually compare different subjects and to visualize changes in network structure as the averaging criterion is varied. 
In the original data, the brain is segmented into 998 regions of interests that cover about $1.5~\textrm{cm}^2$ each and  belong to one of 66 anatomical regions (33 per hemisphere). Here, we do not work with the full network, but rather with regionally aggregated data for the left hemisphere. That is, the data are adjacency matrices in $\R^{33\times 33}$, where $A^{(s)}_{ij}$ is the connection strength between regions $i$ and $j$ for subject $s$. Connections are measured for five different subjects, with two separate measurements for subject A. To compare different subjects and to evaluate whether strong connections are more stable across subjects than weak ones, we perform a thresholding analysis. Our hyperparameter is the quantile of weak connections that we ignore in the comparisons. Thus, $\alpha = 0.1$ corresponds to setting the weakest 10\% of connections to zero. We then binarize the adjacency matrix, $B^{(s)} > A^{(s)}$, such that we have $B^{(s)}_{ij} \in {0,1}$ and compute the Hamming distance between the graphs corresponding to different measurements. We use embeddings in $\R^2$ to represent the data (Fig.~\ref{fig:dti_network}). Surprisingly, the approximate distances between the two measurements of subject A, A1 and A2, are of the same order of magnitude compared to the distances between distinct subjects. This is true across thresholding levels. This is a curious finding, since these two measurements are averaged in \citet{Hagmann2008} before inter-subject averaging is performed. Our findings suggest that these two measurements are not easily distinguishable from other pair comparisons. According to \citet{Hagmann2008}, the correlation between the regional connectivity of A1 and A2 is $r^2 = 0.78$, compared to $r^2 = 0.65$ between distinct subjects. The cMDS visualization suggests that A1 and A2 are not notably more similar than other pairwise comparisons. 

Because of the relatively high noise in the data some form of averaging (over subjects) may be useful to perform structural analyses on the regional network. Due to our findings in the network thresholding analysis we treat measurements A1 and A2 separately. 
One approach to perform averaging is to produce consensus networks out of the individual networks, with the rule that a link is introduced in the 50\% consensus network if and only if it is present in at least 50\% of the individual networks. Thus, we look at the average adjacency matrix $\bar{B}$, where $\bar{B}_{ij}= \left( 1/N \right) \sum_s B^{(s)}_{ij}$ and $B^{(s)} = A^{(s)}>0$. We then threshold it at different levels to produce different consensus networks.  We use the threshold level $\alpha$ as the continuous parameter in cMDS, such that $\bar{B}_{ij}(\alpha) = \bar{B}_{ij}>\alpha$. 
If we take the shortest-path distance on the graph that is defined by $\bar{A}(\alpha)$ as the distance measure between two regions, then changing the threshold level is exactly the same as changing the distance measure, as some links will start disappearing with higher threshold values.  We can therefore apply cMDS to visualize the changes in network structure as the averaging rule is changed. For this example, we implement a weighted version of the algorithm, to mirror the standard Kamada-Kawai layout methods \citep{Kamada1989}. The results are shown in Fig.~\ref{fig:dti_concensus} for regions in the left hemisphere. We also present these results interactively with a web application using the R package {\bf shiny} which is viewable at \href{http://ginagruenhage.shinyapps.io/DTI-App}{http://ginagruenhage.shinyapps.io/DTI-App}. A first (and unsurprising) result is that the network density decreases significantly with the threshold level. That is, as we start requiring higher levels of consistency among the subjects, a lot of connections are rejected. We would like to note, that in this case, integer distances are embedded in a continuous space. However, aspects such centrality are recovered in the visualization: regions with dense connections and high values of centrality and betweenness are placed at the center of the configuration.

Results show that some regions are very stable: for example, the superior frontal cortex (SF) and the superior parietal cortex remain at the core of the network, while the parahippocampal cortex (PARH), the caudal anterior cingulate (CAC) and the temporal cortex (TT) are examples of stable regions at the periphery of the network. What is even more interesting is that some regions are rather unstable and change their role in the network. The postcentral gyrus (PSTS) starts out in the periphery and then moves to the center. Other regions move from the core to the periphery, e.g. the pars orbitalis (PORB), the temporal pole (TP) and the medial orbitofrontal cortex (MOF). Since core-periphery relationships are central to the interpretation of connectome data, it is crucial to know which regions can be reliably called peripheral and others central \citep{Hagmann2008}. cMDS provides this information at a glance. 

\subsection{Hierarchical clustering}
\label{sec:hclust}
We mentioned in the introduction that distance is often computed relative to a certain scale. For spatial or temporal data, scale corresponds to a concrete spatial or temporal window, but there are other ways to obtain a multiscale representation. Hierarchical clustering \citep{Kaufman1990,Hastie2009} is such a technique. We focus here on agglomerative clustering, where the algorithm starts with each observation in one cluster. At each level, the algorithm merges the two closest clusters until only one cluster remains. Thus, there are $N-1$ levels in the hierarchy, which gives a view of the data going from the roughest to the most detailed level. cMDS provides an interesting visualization of the results.

\begin{figure}[h]
%\vspace{0pt}
\centering
\includegraphics[width=\textwidth]{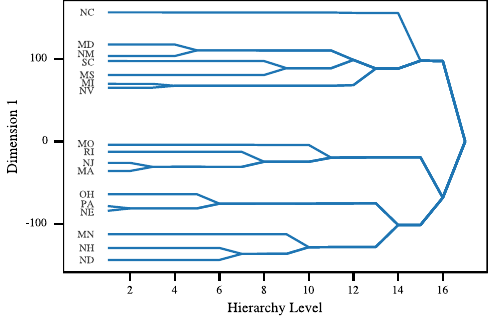}
\caption{Embedding of an hierarchical clustering results for the {\bf USArrests} data set which is publicly available as part of the R \textbf{datasets} package. We selected a sample of 17 states. We ran hierarchical clustering using the centroid of clusters for agglomeration. Then, we built the distance matrix for each level: two datapoints are assigned a small positive distance if they are in the same cluster and the distance between cluster centers otherwise. The cMDS result is a visualization of the tree structure. At each level, the distances between branches represents the distances between cluster centers.}
\label{fig:hclust}
%\vspace{-6pt}
\end{figure}

We first define a distance function at each level of the hierarchy. For each level, we build the distance matrix for the N datapoints as follows: if two datapoints are in the same cluster we assign a very small positive distance. Specifically, we used the absolute value of samples from a Gaussian with zero mean and standard deviation of $0.005$. If two datapoints are not in the same cluster we compute the euclidean distance between the centers of their respective clusters.
Here, we use the publicly available {\bf USArrests} dataset from the R {\bf datasets} package. It contains data on murder arrests, assault arrests, rape arrests and urban population for the different US states in 1973. We picked this dataset because it is used as an example for the R {\bf hclust} function.
We use cMDS to embed these data (Fig.~\ref{fig:hclust}). The result is a tree structure with the property that, at each level of the tree, distances between branches are representative of distances between clusters. This enables an immediate understanding of the hierarchical clustering results. We also developed an interactive visualization based on a 2D embedding, which better captures the potential of cMDS for such applications. We invite readers to have a look at these results (online at \href{http://tinyurl.com/cMDS-demo}{http://tinyurl.com/cMDS-demo}).

\section{Discussion}
We introduced an easy to implement and flexible version of continuous (or dynamic) multi-dimensional scaling, namely cMDS. We showed that cMDS provides a fast and informative way of understanding the data structure by presenting four examples. In a toy example, we compared the results to a method based on $k$-means which gave only an approximate idea of what was going on in the data, while cMDS yielded a very concise representation of the data. With the second example on EU data we showed the effects of changing a weighted metric, putting different weights on two feature classes. Visualization revealed countries whose neighborhood relations are invariant to the change in weights as well as countries whose relative position strongly depends on the weights. In a brain connectivity example, where averaging over subjects is not straightforward, we found a way to compare different subjects giving more and more weight to strong connections. We found that cMDS suggests two measurements of the same subjects to be as different as measurements of two different subjects. In a second analysis we found that the shape of the network changes according to the averaging rule. With cMDS, identifying stable and unstable regions turned out to be straightforward. Finally, we showed how hierarchical clustering can be thought of as a multiscale representation of data, and we used cMDS to visualize how the structure of the data changes across scale.

We only considered small datasets here, for which cMDS is very fast (with a runtime of a few seconds at most). Like other MDS methods, it doesn't scale well with $n$. It hasn't been proved to be better than $\mathcal{O}(n^3)$, though it might be, which depends on how many runs of the outer loop are needed. Practically, the runtime remains reasonable with a few hundred datapoints, and straightforward extensions for larger datasets are possible. One idea is to embed a subset of landmark points, as in landmark MDS \citep{Silva2003}. 
Another is to use sparse weighting matrices, tying each datapoint to a random subset of neighbours. 

By visual inspection, cMDS immediately reveals qualitative structures in the neighborhood dynamics for various datasets. Furthermore, quantitative analyses are possible. For example, performing clustering on cMDS output can yield results that are robust to changes in the distance measure. We leave these extensions to future work. 

\section{Conclusion}
With cMDS, we address a fundamental problem in pattern recognition and machine learning: the initial choice of a distance metric. This is a hidden assumption in various methods. We argue that this choice should be addressed explicitly. Our suggestion is to use continuous MDS techniques, which visualize the dynamics that (so far) arbitrary choices in distance functions introduce in data. cMDS can deal with numerous sources of arbitrariness in the distance metric, examples of which are varying scale or weighting. We show that interesting and important dynamics, such as invariance and declustering, are readily revealed by cMDS. Finally, we provide a cMDS algorithm that is straightforward to implement, use and extend. 

\section*{Acknowledgments}
This work was partially supported by the Deutsche Forschungsgemeinschaft (GRK1589/1).
\newpage

\section*{References}

\bibliography{ggruenhage}

\begin{thebibliography}{62}
\providecommand{\natexlab}[1]{#1}
\providecommand{\url}[1]{\texttt{#1}}
\providecommand{\href}[2]{#2}
\providecommand{\path}[1]{#1}
\providecommand{\DOIprefix}{doi:}
\providecommand{\ArXivprefix}{arXiv:}
\providecommand{\URLprefix}{URL: }
\providecommand{\Pubmedprefix}{pmid:}
\providecommand{\doi}[1]{\href{http://dx.doi.org/#1}{\path{#1}}}
\providecommand{\Pubmed}[1]{\href{pmid:#1}{\path{#1}}}
\providecommand{\BIBand}{and}
\providecommand{\bibinfo}[2]{#2}
\ifx\xfnm\undefined \def\xfnm[#1]{\unskip,\space#1}\fi
\makeatletter\def\@biblabel#1{#1.}\makeatother
%Type = Article
\bibitem[{Carlsson(2009)}]{Carlsson2009}
\bibinfo{author}{Carlsson\xfnm[ G.]}.
\newblock \bibinfo{title}{{Topology and data}}.
\newblock \emph{\bibinfo{journal}{Bulletin of the American Mathematical
  Society}}
  \bibinfo{year}{2009};\bibinfo{volume}{46}(\bibinfo{number}{2}):\bibinfo{pages}{255--308}.
%Type = Book
\bibitem[{Sch{\"{o}}lkopf and Smola(2002)}]{Scholkopf2002}
\bibinfo{author}{Sch{\"{o}}lkopf\xfnm[ B.]}, \bibinfo{author}{Smola\xfnm[
  A.J.]}.
\newblock \bibinfo{title}{{Learning with kernels: support vector machines,
  regularization, optimization and beyond}}.
\newblock \bibinfo{publisher}{the MIT Press}; \bibinfo{year}{2002}.
%Type = Article
\bibitem[{Lum et~al.(2013)Lum, Singh, Lehman, Ishkanov, Vejdemo-Johansson,
  Alagappan, Carlsson and Carlsson}]{Lum2013}
\bibinfo{author}{Lum\xfnm[ P.Y.]}, \bibinfo{author}{Singh\xfnm[ G.]},
  \bibinfo{author}{Lehman\xfnm[ A.]}, \bibinfo{author}{Ishkanov\xfnm[ T.]},
  \bibinfo{author}{Vejdemo-Johansson\xfnm[ M.]},
  \bibinfo{author}{Alagappan\xfnm[ M.]}, \bibinfo{author}{Carlsson\xfnm[ J.]},
  \bibinfo{author}{Carlsson\xfnm[ G.]}.
\newblock \bibinfo{title}{{Extracting insights from the shape of complex data
  using topology}}.
\newblock \emph{\bibinfo{journal}{Scientific reports}}
  \bibinfo{year}{2013};\bibinfo{volume}{3}.
%Type = Article
\bibitem[{Buja and Swayne(2002)}]{Buja2002}
\bibinfo{author}{Buja\xfnm[ A.]}, \bibinfo{author}{Swayne\xfnm[ D.F.]}.
\newblock \bibinfo{title}{{Visualization methodology for multidimensional
  scaling}}.
\newblock \emph{\bibinfo{journal}{Journal of Classification}}
  \bibinfo{year}{2002};\bibinfo{volume}{19}(\bibinfo{number}{1}):\bibinfo{pages}{7--43}.
%Type = Article
\bibitem[{Buja et~al.(2008)Buja, Swayne, Littman, Dean, Hofmann and
  Chen}]{Buja2008}
\bibinfo{author}{Buja\xfnm[ A.]}, \bibinfo{author}{Swayne\xfnm[ D.F.]},
  \bibinfo{author}{Littman\xfnm[ M.L.]}, \bibinfo{author}{Dean\xfnm[ N.]},
  \bibinfo{author}{Hofmann\xfnm[ H.]}, \bibinfo{author}{Chen\xfnm[ L.]}.
\newblock \bibinfo{title}{{Data Visualization With Multidimensional Scaling}}.
\newblock \emph{\bibinfo{journal}{Journal of Computational and Graphical
  Statistics}}
  \bibinfo{year}{2008};\bibinfo{volume}{17}(\bibinfo{number}{2}):\bibinfo{pages}{444--472}.
%Type = Article
\bibitem[{Torgerson(1952)}]{Torgerson1952}
\bibinfo{author}{Torgerson\xfnm[ W.]}.
\newblock \bibinfo{title}{{Multidimensional scaling: I. Theory and method}}.
\newblock \emph{\bibinfo{journal}{Psychometrika}}
  \bibinfo{year}{1952};\bibinfo{volume}{17}(\bibinfo{number}{4}):\bibinfo{pages}{401--419}.
%Type = Book
\bibitem[{Torgerson(1958)}]{Torgerson1958}
\bibinfo{author}{Torgerson\xfnm[ W.S.]}.
\newblock \bibinfo{title}{{Theory and methods of scaling.}}
\newblock \bibinfo{publisher}{Wiley}; \bibinfo{year}{1958}.
%Type = Article
\bibitem[{Gower(1966)}]{Gower1966}
\bibinfo{author}{Gower\xfnm[ J.C.]}.
\newblock \bibinfo{title}{{Some distance properties of latent root and vector
  methods used in multivariate analysis}}.
\newblock \emph{\bibinfo{journal}{Biometrika}}
  \bibinfo{year}{1966};\bibinfo{volume}{53}(\bibinfo{number}{3-4}):\bibinfo{pages}{325--338}.
%Type = Article
\bibitem[{Kruskal(1964)}]{Kruskal1964}
\bibinfo{author}{Kruskal\xfnm[ J.B.]}.
\newblock \bibinfo{title}{{Multidimensional scaling by optimizing goodness of
  fit to a nonmetric hypothesis}}.
\newblock \emph{\bibinfo{journal}{Psychometrika}}
  \bibinfo{year}{1964};\bibinfo{volume}{29}(\bibinfo{number}{1}):\bibinfo{pages}{1--27}.
%Type = Article
\bibitem[{Shepard(1962)}]{Shepard1962}
\bibinfo{author}{Shepard\xfnm[ R.N.]}.
\newblock \bibinfo{title}{{The analysis of proximities: Multidimensional
  scaling with an unknown distance function. I.}}
\newblock \emph{\bibinfo{journal}{Psychometrika}}
  \bibinfo{year}{1962};\bibinfo{volume}{27}(\bibinfo{number}{2}):\bibinfo{pages}{125--140}.
%Type = Article
\bibitem[{Ramsay(1977)}]{Ramsay1977}
\bibinfo{author}{Ramsay\xfnm[ J.O.]}.
\newblock \bibinfo{title}{{Maximum likelihood estimation in multidimensional
  scaling}}.
\newblock \emph{\bibinfo{journal}{Psychometrika}}
  \bibinfo{year}{1977};\bibinfo{volume}{42}(\bibinfo{number}{2}):\bibinfo{pages}{241--266}.
%Type = Article
\bibitem[{Ramsay(1978{\natexlab{a}})}]{Ramsay1978}
\bibinfo{author}{Ramsay\xfnm[ J.O.]}.
\newblock \bibinfo{title}{{Confidence regions for multidimensional scaling
  analysis}}.
\newblock \emph{\bibinfo{journal}{Psychometrika}}
  \bibinfo{year}{1978}{\natexlab{a}};\bibinfo{volume}{43}(\bibinfo{number}{2}):\bibinfo{pages}{145--160}.
%Type = Book
\bibitem[{Ramsay(1978{\natexlab{b}})}]{Ramsay1978b}
\bibinfo{author}{Ramsay\xfnm[ J.O.]}.
\newblock \bibinfo{title}{{Multiscale: Four Programs for Multidimensional
  Scaling by the Method of Maximum Likelihood.[user's Guide]}}.
\newblock \bibinfo{publisher}{National Educational Resources};
  \bibinfo{year}{1978}{\natexlab{b}}.
%Type = Article
\bibitem[{Sammon(1969)}]{Sammon1969}
\bibinfo{author}{Sammon\xfnm[ J.W.]}.
\newblock \bibinfo{title}{{A nonlinear mapping for data structure analysis}}.
\newblock \emph{\bibinfo{journal}{Computers, IEEE Transactions on}}
  \bibinfo{year}{1969};\bibinfo{volume}{100}(\bibinfo{number}{5}):\bibinfo{pages}{401--409}.
%Type = Article
\bibitem[{McGee(1966)}]{McGee1966}
\bibinfo{author}{McGee\xfnm[ V.E.]}.
\newblock \bibinfo{title}{{The multidimensional analysis of
  'elastic'distances}}.
\newblock \emph{\bibinfo{journal}{British Journal of Mathematical and
  Statistical Psychology}}
  \bibinfo{year}{1966};\bibinfo{volume}{19}(\bibinfo{number}{2}):\bibinfo{pages}{181--196}.
%Type = Book
\bibitem[{Borg and Groenen(2005)}]{Borg2005}
\bibinfo{author}{Borg\xfnm[ I.]}, \bibinfo{author}{Groenen\xfnm[ P.J.F.]}.
\newblock \bibinfo{title}{{Modern Multidimensional Scaling: Theory and
  Applications (Springer Series in Statistics)}}.
\newblock \bibinfo{edition}{Second} ed.; \bibinfo{publisher}{Springer};
  \bibinfo{year}{2005}.
%Type = Article
\bibitem[{Chen and Buja(2009)}]{Chen2009}
\bibinfo{author}{Chen\xfnm[ L.]}, \bibinfo{author}{Buja\xfnm[ A.]}.
\newblock \bibinfo{title}{{Local multidimensional scaling for nonlinear
  dimension reduction, graph drawing, and proximity analysis}}.
\newblock \emph{\bibinfo{journal}{Journal of the American Statistical
  Association}}
  \bibinfo{year}{2009};\bibinfo{volume}{104}(\bibinfo{number}{485}):\bibinfo{pages}{209--219}.
%Type = Article
\bibitem[{Tenenbaum et~al.(2000)Tenenbaum, de~Silva and
  Langford}]{Tenenbaum2000}
\bibinfo{author}{Tenenbaum\xfnm[ J.B.]}, \bibinfo{author}{de~Silva\xfnm[ V.]},
  \bibinfo{author}{Langford\xfnm[ J.C.]}.
\newblock \bibinfo{title}{{A Global Geometric Framework for Nonlinear
  Dimensionality Reduction}}.
\newblock \emph{\bibinfo{journal}{Science}}
  \bibinfo{year}{2000};\bibinfo{volume}{290}(\bibinfo{number}{5500}).
%Type = Article
\bibitem[{Guttman(1968)}]{Guttman1968}
\bibinfo{author}{Guttman\xfnm[ L.]}.
\newblock \bibinfo{title}{{A general nonmetric technique for finding the
  smallest coordinate space for a configuration of points}}.
\newblock \emph{\bibinfo{journal}{Psychometrika}}
  \bibinfo{year}{1968};\bibinfo{volume}{33}(\bibinfo{number}{4}):\bibinfo{pages}{469--506}.
%Type = Incollection
\bibitem[{De~Leeuw(1977)}]{DeLeeuw1977}
\bibinfo{author}{De~Leeuw\xfnm[ J.]}.
\newblock \bibinfo{title}{{Applications of Convex Analysis to Multidimensional
  Scaling}}.
\newblock In: \bibinfo{editor}{Barra\xfnm[ J.R.]},
  \bibinfo{editor}{Brodeau\xfnm[ F.]}, \bibinfo{editor}{Romier\xfnm[ G.]},
  \bibinfo{editor}{van Cutsem\xfnm[ B.]}, eds. \emph{\bibinfo{booktitle}{Recent
  developments in statistics}}. \bibinfo{address}{Amsterdam, The Netherlands}:
  \bibinfo{publisher}{North-Holland}; \bibinfo{year}{1977}:\unskip
  \bibinfo{pages}{133--145}.
%Type = Article
\bibitem[{De~Leeuw and Heiser(1977)}]{DeLeeuw1977b}
\bibinfo{author}{De~Leeuw\xfnm[ J.]}, \bibinfo{author}{Heiser\xfnm[ W.J.]}.
\newblock \bibinfo{title}{{Convergence of correction matrix algorithms for
  multidimensional scaling}}.
\newblock \emph{\bibinfo{journal}{Geometric representations of relational
  data}} \bibinfo{year}{1977};:\bibinfo{pages}{735--752}.
%Type = Article
\bibitem[{De~Leeuw(1988)}]{DeLeeuw1988}
\bibinfo{author}{De~Leeuw\xfnm[ J.]}.
\newblock \bibinfo{title}{{Convergence of the majorization method for
  multidimensional scaling}}.
\newblock \emph{\bibinfo{journal}{Journal of Classification}}
  \bibinfo{year}{1988};\bibinfo{volume}{5}(\bibinfo{number}{2}):\bibinfo{pages}{163--180}.
%Type = Article
\bibitem[{Kamada and Kawai(1989)}]{Kamada1989}
\bibinfo{author}{Kamada\xfnm[ T.]}, \bibinfo{author}{Kawai\xfnm[ S.]}.
\newblock \bibinfo{title}{{An algorithm for drawing general undirected
  graphs}}.
\newblock \emph{\bibinfo{journal}{Information processing letters}}
  \bibinfo{year}{1989};\bibinfo{volume}{31}(\bibinfo{number}{1}):\bibinfo{pages}{7--15}.
%Type = Incollection
\bibitem[{Gansner et~al.(2005)Gansner, Koren and North}]{Gansner2005}
\bibinfo{author}{Gansner\xfnm[ E.]}, \bibinfo{author}{Koren\xfnm[ Y.]},
  \bibinfo{author}{North\xfnm[ S.]}.
\newblock \bibinfo{title}{{Graph Drawing by Stress Majorization}}.
\newblock In: \bibinfo{editor}{Pach\xfnm[ J.]}, ed.
  \emph{\bibinfo{booktitle}{Graph Drawing}}; vol. \bibinfo{volume}{3383} of
  \emph{\bibinfo{series}{Lecture Notes in Computer Science}}.
  \bibinfo{publisher}{Springer Berlin Heidelberg}; \bibinfo{year}{2005}:\unskip
  \bibinfo{pages}{239--250}.
%Type = Article
\bibitem[{Misue et~al.(1995)Misue, Eades, Lai and Sugiyama}]{Misue1995}
\bibinfo{author}{Misue\xfnm[ K.]}, \bibinfo{author}{Eades\xfnm[ P.]},
  \bibinfo{author}{Lai\xfnm[ W.]}, \bibinfo{author}{Sugiyama\xfnm[ K.]}.
\newblock \bibinfo{title}{{Layout Adjustment and the Mental Map}}.
\newblock \emph{\bibinfo{journal}{Journal of Visual Languages \& Computing}}
  \bibinfo{year}{1995};\bibinfo{volume}{6}(\bibinfo{number}{2}):\bibinfo{pages}{183--210}.
%Type = Inproceedings
\bibitem[{B\"{o}hringer and Paulisch(1990)}]{Boehringer1990}
\bibinfo{author}{B\"{o}hringer\xfnm[ K.F.]}, \bibinfo{author}{Paulisch\xfnm[
  F.N.]}.
\newblock \bibinfo{title}{{Using constraints to achieve stability in automatic
  graph layout algorithms}}.
\newblock In: \emph{\bibinfo{booktitle}{Proceedings of the SIGCHI Conference on
  Human Factors in Computing Systems}}. CHI '90; \bibinfo{address}{New York,
  NY, USA}: \bibinfo{publisher}{ACM}; \bibinfo{year}{1990}:\unskip
  \bibinfo{pages}{43--51}.
%Type = Incollection
\bibitem[{North(1996)}]{North1996}
\bibinfo{author}{North\xfnm[ S.]}.
\newblock \bibinfo{title}{{Incremental layout in DynaDAG}}.
\newblock In: \bibinfo{editor}{Brandenburg\xfnm[ F.]}, ed.
  \emph{\bibinfo{booktitle}{Graph Drawing}}; vol. \bibinfo{volume}{1027} of
  \emph{\bibinfo{series}{Lecture Notes in Computer Science}}.
  \bibinfo{publisher}{Springer Berlin Heidelberg}; \bibinfo{year}{1996}:\unskip
  \bibinfo{pages}{409--418}.
%Type = Incollection
\bibitem[{Brandes and Wagner(1997)}]{Brandes1997}
\bibinfo{author}{Brandes\xfnm[ U.]}, \bibinfo{author}{Wagner\xfnm[ D.]}.
\newblock \bibinfo{title}{{A bayesian paradigm for dynamic graph layout}}.
\newblock In: \bibinfo{editor}{DiBattista\xfnm[ G.]}, ed.
  \emph{\bibinfo{booktitle}{Graph Drawing}}; vol. \bibinfo{volume}{1353} of
  \emph{\bibinfo{series}{Lecture Notes in Computer Science}}.
  \bibinfo{publisher}{Springer Berlin Heidelberg}; \bibinfo{year}{1997}:\unskip
  \bibinfo{pages}{236--247}.
%Type = Article
\bibitem[{Brandes and Corman(2003)}]{Brandes2003}
\bibinfo{author}{Brandes\xfnm[ U.]}, \bibinfo{author}{Corman\xfnm[ S.R.]}.
\newblock \bibinfo{title}{{Visual Unrolling of Network Evolution and the
  Analysis of Dynamic Discourse{\dag}}}.
\newblock \emph{\bibinfo{journal}{Information Visualization}}
  \bibinfo{year}{2003};\bibinfo{volume}{2}(\bibinfo{number}{1}):\bibinfo{pages}{40--50}.
%Type = Article
\bibitem[{Moody et~al.(2005)Moody, McFarland and Bender‐deMoll}]{Moody2005}
\bibinfo{author}{Moody\xfnm[ J.]}, \bibinfo{author}{McFarland\xfnm[ D.]},
  \bibinfo{author}{Bender‐deMoll\xfnm[ S.]}.
\newblock \bibinfo{title}{{Dynamic Network Visualization}}.
\newblock \emph{\bibinfo{journal}{American Journal of Sociology}}
  \bibinfo{year}{2005};\bibinfo{volume}{110}(\bibinfo{number}{4}):\bibinfo{pages}{1206--1241}.
%Type = Incollection
\bibitem[{Diehl and G\"{o}rg(2002)}]{Diehl2002}
\bibinfo{author}{Diehl\xfnm[ S.]}, \bibinfo{author}{G\"{o}rg\xfnm[ C.]}.
\newblock \bibinfo{title}{{Graphs, They Are Changing}}.
\newblock In: \bibinfo{editor}{Goodrich\xfnm[ M.]},
  \bibinfo{editor}{Kobourov\xfnm[ S.]}, eds. \emph{\bibinfo{booktitle}{Graph
  Drawing}}; vol. \bibinfo{volume}{2528} of \emph{\bibinfo{series}{Lecture
  Notes in Computer Science}}. \bibinfo{publisher}{Springer Berlin Heidelberg};
  \bibinfo{year}{2002}:\unskip \bibinfo{pages}{23--31}.
%Type = Article
\bibitem[{Frishman and Tal(2008)}]{Frishman2008}
\bibinfo{author}{Frishman\xfnm[ Y.]}, \bibinfo{author}{Tal\xfnm[ A.]}.
\newblock \bibinfo{title}{{Online Dynamic Graph Drawing}}.
\newblock \emph{\bibinfo{journal}{Visualization and Computer Graphics, IEEE
  Transactions on}}
  \bibinfo{year}{2008};\bibinfo{volume}{14}(\bibinfo{number}{4}):\bibinfo{pages}{727--740}.
%Type = Inproceedings
\bibitem[{Erten et~al.(2004{\natexlab{a}})Erten, Harding, Kobourov, Wampler and
  Yee}]{Erten2004}
\bibinfo{author}{Erten\xfnm[ C.]}, \bibinfo{author}{Harding\xfnm[ P.J.]},
  \bibinfo{author}{Kobourov\xfnm[ S.G.]}, \bibinfo{author}{Wampler\xfnm[ K.]},
  \bibinfo{author}{Yee\xfnm[ G.]}.
\newblock \bibinfo{title}{{Exploring the computing literature using temporal
  graph visualization}}.
\newblock In: \emph{\bibinfo{booktitle}{Visualization and Data Analysis 2004.
  Edited by Erbacher, Robert F.; Chen, Philip C.; Roberts, Jonathan C.;
  Gr\"ohn, Matti T.; B{\"o}rner, Katy. Proceedings of the SPIE, Volume 5295,
  pp. 45-56 (2004).}}; vol. \bibinfo{volume}{5295}.
  \bibinfo{year}{2004}{\natexlab{a}}:\unskip \bibinfo{pages}{45--56}.
%Type = Incollection
\bibitem[{Erten et~al.(2004{\natexlab{b}})Erten, Kobourov, Le and
  Navabi}]{Erten2004b}
\bibinfo{author}{Erten\xfnm[ C.]}, \bibinfo{author}{Kobourov\xfnm[ S.]},
  \bibinfo{author}{Le\xfnm[ V.]}, \bibinfo{author}{Navabi\xfnm[ A.]}.
\newblock \bibinfo{title}{{Simultaneous Graph Drawing: Layout Algorithms and
  Visualization Schemes}}.
\newblock In: \bibinfo{editor}{Liotta\xfnm[ G.]}, ed.
  \emph{\bibinfo{booktitle}{Graph Drawing}}; vol. \bibinfo{volume}{2912} of
  \emph{\bibinfo{series}{Lecture Notes in Computer Science}};
  chap.~\bibinfo{chapter}{41}. \bibinfo{address}{Berlin, Heidelberg}:
  \bibinfo{publisher}{Springer Berlin Heidelberg};
  \bibinfo{year}{2004}{\natexlab{b}}:\unskip \bibinfo{pages}{437--449}.
%Type = Inproceedings
\bibitem[{Dwyer et~al.(2006)Dwyer, Hong, Kosch\"{u}tzki, Schreiber and
  Xu}]{Dwyer2006}
\bibinfo{author}{Dwyer\xfnm[ T.]}, \bibinfo{author}{Hong\xfnm[ S.H.]},
  \bibinfo{author}{Kosch\"{u}tzki\xfnm[ D.]}, \bibinfo{author}{Schreiber\xfnm[
  F.]}, \bibinfo{author}{Xu\xfnm[ K.]}.
\newblock \bibinfo{title}{{Visual analysis of network centralities}}.
\newblock In: \emph{\bibinfo{booktitle}{Proceedings of the 2006 Asia-Pacific
  Symposium on Information Visualisation - Volume 60}}. APVis '06;
  \bibinfo{address}{Darlinghurst, Australia, Australia}:
  \bibinfo{publisher}{Australian Computer Society, Inc.};
  \bibinfo{year}{2006}:\unskip \bibinfo{pages}{189--197}.
%Type = Book
\bibitem[{Baur and Schank(2008)}]{Baur2008}
\bibinfo{author}{Baur\xfnm[ M.]}, \bibinfo{author}{Schank\xfnm[ T.]}.
\newblock \bibinfo{title}{{Dynamic graph drawing in visone}}.
\newblock \bibinfo{publisher}{Citeseer}; \bibinfo{year}{2008}.
%Type = Incollection
\bibitem[{Brandes and Mader(2012)}]{Brandes2012a}
\bibinfo{author}{Brandes\xfnm[ U.]}, \bibinfo{author}{Mader\xfnm[ M.]}.
\newblock \bibinfo{title}{{A Quantitative Comparison of Stress-Minimization
  Approaches for Offline Dynamic Graph Drawing}}.
\newblock In: \bibinfo{editor}{Kreveld\xfnm[ M.]},
  \bibinfo{editor}{Speckmann\xfnm[ B.]}, eds. \emph{\bibinfo{booktitle}{Graph
  Drawing}}; vol. \bibinfo{volume}{7034} of \emph{\bibinfo{series}{Lecture
  Notes in Computer Science}}. \bibinfo{publisher}{Springer Berlin Heidelberg};
  \bibinfo{year}{2012}:\unskip \bibinfo{pages}{99--110}.
%Type = Article
\bibitem[{Xu et~al.(2013)Xu, Kliger and Hero}]{Xu2013}
\bibinfo{author}{Xu\xfnm[ K.]}, \bibinfo{author}{Kliger\xfnm[ M.]},
  \bibinfo{author}{Hero\xfnm[ A.]}.
\newblock \bibinfo{title}{{A regularized graph layout framework for dynamic
  network visualization}}.
\newblock \emph{\bibinfo{journal}{Data Mining and Knowledge Discovery}}
  \bibinfo{year}{2013};\bibinfo{volume}{27}(\bibinfo{number}{1}):\bibinfo{pages}{84--116}.
%Type = Article
\bibitem[{Sarkar and Moore(2005)}]{Sarkar2005}
\bibinfo{author}{Sarkar\xfnm[ P.]}, \bibinfo{author}{Moore\xfnm[ A.W.]}.
\newblock \bibinfo{title}{{Dynamic social network analysis using latent space
  models}}.
\newblock \emph{\bibinfo{journal}{SIGKDD Explor Newsl}}
  \bibinfo{year}{2005};\bibinfo{volume}{7}(\bibinfo{number}{2}):\bibinfo{pages}{31--40}.
%Type = Article
\bibitem[{Mizuta(2003)}]{Masahiro2003}
\bibinfo{author}{Mizuta\xfnm[ M.]}.
\newblock \bibinfo{title}{Multidimensional scaling for dissimilarity functions
  with continuous argument(functional data analysis)}.
\newblock \emph{\bibinfo{journal}{Journal of the Japanese Society of
  Computational Statistics}}
  \bibinfo{year}{2003};\bibinfo{volume}{15}(\bibinfo{number}{2}):\bibinfo{pages}{327--333}.
%Type = Article
\bibitem[{Yuille and Rangarajan(2003)}]{Yuille2003}
\bibinfo{author}{Yuille\xfnm[ A.L.]}, \bibinfo{author}{Rangarajan\xfnm[ A.]}.
\newblock \bibinfo{title}{{The Concave-convex Procedure}}.
\newblock \emph{\bibinfo{journal}{Neural Comput}}
  \bibinfo{year}{2003};\bibinfo{volume}{15}(\bibinfo{number}{4}):\bibinfo{pages}{915--936}.
%Type = Incollection
\bibitem[{Lanckriet and Sriperumbudur(2009)}]{Sriperumbudur2009}
\bibinfo{author}{Lanckriet\xfnm[ G.R.]}, \bibinfo{author}{Sriperumbudur\xfnm[
  B.K.]}.
\newblock \bibinfo{title}{{On the Convergence of the Concave-Convex
  Procedure}}.
\newblock In: \bibinfo{editor}{Bengio\xfnm[ Y.]},
  \bibinfo{editor}{Schuurmans\xfnm[ D.]}, \bibinfo{editor}{Lafferty\xfnm[
  J.D.]}, \bibinfo{editor}{Williams\xfnm[ C.K.I.]},
  \bibinfo{editor}{Culotta\xfnm[ A.]}, eds. \emph{\bibinfo{booktitle}{Advances
  in Neural Information Processing Systems 22}}. \bibinfo{publisher}{Curran
  Associates, Inc.}; \bibinfo{year}{2009}:\unskip \bibinfo{pages}{1759--1767}.
%Type = Inproceedings
\bibitem[{Yen et~al.(2012)Yen, Peng, Wang and Lin}]{Yen2012}
\bibinfo{author}{Yen\xfnm[ I.E.]}, \bibinfo{author}{Peng\xfnm[ N.]},
  \bibinfo{author}{Wang\xfnm[ P.]}, \bibinfo{author}{Lin\xfnm[ S.]}.
\newblock \bibinfo{title}{{On convergence rate of concave-convex procedure}}.
\newblock In: \emph{\bibinfo{booktitle}{Proceedings of the NIPS 2012
  Optimization Workshop}}. \bibinfo{year}{2012}:\unskip.
%Type = Book
\bibitem[{Cook and Swayne(2007)}]{Cook2007}
\bibinfo{author}{Cook\xfnm[ D.]}, \bibinfo{author}{Swayne\xfnm[ D.F.]}.
\newblock \bibinfo{title}{{Interactive and dynamic graphics for data analysis:
  with R and GGobi}}.
\newblock \bibinfo{publisher}{Springer}; \bibinfo{year}{2007}.
%Type = Book
\bibitem[{Ramsay and Silverman(1997)}]{Ramsay1997}
\bibinfo{author}{Ramsay\xfnm[ J.O.]}, \bibinfo{author}{Silverman\xfnm[ B.W.]}.
\newblock \bibinfo{title}{{Functional Data Analysis}}.
\newblock \bibinfo{address}{New York}: \bibinfo{publisher}{Springer Series in
  Statistics}; \bibinfo{year}{1997}.
%Type = Article
\bibitem[{Kaski and Peltonen(2011)}]{Kaski2011}
\bibinfo{author}{Kaski\xfnm[ S.]}, \bibinfo{author}{Peltonen\xfnm[ J.]}.
\newblock \bibinfo{title}{{Dimensionality Reduction for Data Visualization
  [Applications Corner]}}.
\newblock \emph{\bibinfo{journal}{Signal Processing Magazine, IEEE}}
  \bibinfo{year}{2011};\bibinfo{volume}{28}(\bibinfo{number}{2}):\bibinfo{pages}{100--104}.
%Type = Article
\bibitem[{Mokbel et~al.(2013)Mokbel, Lueks, Gisbrecht and Hammer}]{Mokbel2013}
\bibinfo{author}{Mokbel\xfnm[ B.]}, \bibinfo{author}{Lueks\xfnm[ W.]},
  \bibinfo{author}{Gisbrecht\xfnm[ A.]}, \bibinfo{author}{Hammer\xfnm[ B.]}.
\newblock \bibinfo{title}{{Visualizing the quality of dimensionality
  reduction}}.
\newblock \emph{\bibinfo{journal}{Neurocomputing}}
  \bibinfo{year}{2013};\bibinfo{volume}{112}:\bibinfo{pages}{109--123}.
%Type = Article
\bibitem[{Rockafellar(1970)}]{Rockafellar1970}
\bibinfo{author}{Rockafellar\xfnm[ R.T.]}.
\newblock \bibinfo{title}{Convex analysis (princeton mathematical series)}.
\newblock \emph{\bibinfo{journal}{Princeton University Press}}
  \bibinfo{year}{1970};\bibinfo{volume}{46}:\bibinfo{pages}{49}.
%Type = Inproceedings
\bibitem[{Agarwal et~al.(2010)Agarwal, Phillips and
  Venkatasubramanian}]{Agarwal2010}
\bibinfo{author}{Agarwal\xfnm[ A.]}, \bibinfo{author}{Phillips\xfnm[ J.M.]},
  \bibinfo{author}{Venkatasubramanian\xfnm[ S.]}.
\newblock \bibinfo{title}{{Universal multi-dimensional scaling}}.
\newblock In: \emph{\bibinfo{booktitle}{Proceedings of the 16th ACM SIGKDD
  international conference on Knowledge discovery and data mining}}.
  \bibinfo{organization}{ACM}; \bibinfo{year}{2010}:\unskip
  \bibinfo{pages}{1149--1158}.
%Type = Incollection
\bibitem[{Leeuw(1994)}]{DeLeeuw1994}
\bibinfo{author}{Leeuw\xfnm[ J.]}.
\newblock \bibinfo{title}{{Block-relaxation Algorithms in Statistics}}.
\newblock In: \bibinfo{editor}{Bock\xfnm[ H.H.]}, \bibinfo{editor}{Lenski\xfnm[
  W.]}, \bibinfo{editor}{Richter\xfnm[ M.]}, eds.
  \emph{\bibinfo{booktitle}{Information Systems and Data Analysis}}. Studies in
  Classification, Data Analysis, and Knowledge Organization;
  \bibinfo{publisher}{Springer Berlin Heidelberg}; \bibinfo{year}{1994}:\unskip
  \bibinfo{pages}{308--324}.
%Type = Article
\bibitem[{Hunter and Lange(2004)}]{Hunter2004}
\bibinfo{author}{Hunter\xfnm[ D.R.]}, \bibinfo{author}{Lange\xfnm[ K.]}.
\newblock \bibinfo{title}{{A tutorial on MM algorithms}}.
\newblock \emph{\bibinfo{journal}{The American Statistician}}
  \bibinfo{year}{2004};\bibinfo{volume}{58}(\bibinfo{number}{1}):\bibinfo{pages}{30--37}.
%Type = Article
\bibitem[{Tseng and Yun(2009)}]{Tseng2009}
\bibinfo{author}{Tseng\xfnm[ P.]}, \bibinfo{author}{Yun\xfnm[ S.]}.
\newblock \bibinfo{title}{{A coordinate gradient descent method for nonsmooth
  separable minimization}}.
\newblock \emph{\bibinfo{journal}{Mathematical Programming}}
  \bibinfo{year}{2009};\bibinfo{volume}{117}(\bibinfo{number}{1-2}):\bibinfo{pages}{387--423}.
%Type = Article
\bibitem[{Razaviyayn et~al.(2013)Razaviyayn, Hong and Luo}]{Razaviyayn2013}
\bibinfo{author}{Razaviyayn\xfnm[ M.]}, \bibinfo{author}{Hong\xfnm[ M.]},
  \bibinfo{author}{Luo\xfnm[ Z.Q.]}.
\newblock \bibinfo{title}{{A unified convergence analysis of block successive
  minimization methods for nonsmooth optimization}}.
\newblock \emph{\bibinfo{journal}{SIAM Journal on Optimization}}
  \bibinfo{year}{2013};\bibinfo{volume}{23}(\bibinfo{number}{2}):\bibinfo{pages}{1126--1153}.
%Type = Article
\bibitem[{Wright(2015)}]{Wright2015}
\bibinfo{author}{Wright\xfnm[ S.]}.
\newblock \bibinfo{title}{Coordinate descent algorithms}.
\newblock \emph{\bibinfo{journal}{Mathematical Programming}}
  \bibinfo{year}{2015};\bibinfo{volume}{151}(\bibinfo{number}{1}):\bibinfo{pages}{3--34}.
%Type = Article
\bibitem[{Powell(1973)}]{Powell1973}
\bibinfo{author}{Powell\xfnm[ M.J.]}.
\newblock \bibinfo{title}{On search directions for minimization algorithms}.
\newblock \emph{\bibinfo{journal}{Mathematical Programming}}
  \bibinfo{year}{1973};\bibinfo{volume}{4}(\bibinfo{number}{1}):\bibinfo{pages}{193--201}.
%Type = Article
\bibitem[{De~Leeuw and Patrick(2009)}]{DeLeeuw2009}
\bibinfo{author}{De~Leeuw\xfnm[ J.]}, \bibinfo{author}{Patrick\xfnm[ M.]}.
\newblock \bibinfo{title}{{Multidimensional scaling using majorization: SMACOF
  in R}}.
\newblock \emph{\bibinfo{journal}{Journal of Statistical Software}}
  \bibinfo{year}{2009};\bibinfo{volume}{31}(\bibinfo{number}{3}):\bibinfo{pages}{1--30}.
%Type = Article
\bibitem[{Hagmann et~al.(2008)Hagmann, Cammoun, Gigandet, Meuli, Honey, Wedeen
  and Sporns}]{Hagmann2008}
\bibinfo{author}{Hagmann\xfnm[ P.]}, \bibinfo{author}{Cammoun\xfnm[ L.]},
  \bibinfo{author}{Gigandet\xfnm[ X.]}, \bibinfo{author}{Meuli\xfnm[ R.]},
  \bibinfo{author}{Honey\xfnm[ C.J.]}, \bibinfo{author}{Wedeen\xfnm[ V.J.]},
  \bibinfo{author}{Sporns\xfnm[ O.]}.
\newblock \bibinfo{title}{{Mapping the Structural Core of Human Cerebral
  Cortex}}.
\newblock \emph{\bibinfo{journal}{PLoS Biol}}
  \bibinfo{year}{2008};\bibinfo{volume}{6}(\bibinfo{number}{7}):\bibinfo{pages}{e159+}.
%Type = Article
\bibitem[{Sporns et~al.(2005)Sporns, Tononi and K\"{o}tter}]{Koetter2005}
\bibinfo{author}{Sporns\xfnm[ O.]}, \bibinfo{author}{Tononi\xfnm[ G.]},
  \bibinfo{author}{K\"{o}tter\xfnm[ R.]}.
\newblock \bibinfo{title}{{The human connectome: A structural description of
  the human brain.}}
\newblock \emph{\bibinfo{journal}{PLoS computational biology}}
  \bibinfo{year}{2005};\bibinfo{volume}{1}(\bibinfo{number}{4}):\bibinfo{pages}{e42+}.
%Type = Article
\bibitem[{Hagmann et~al.(2003)Hagmann, Thiran, Jonasson, Vandergheynst, Clarke,
  Maeder and Meuli}]{Hagmann2003}
\bibinfo{author}{Hagmann\xfnm[ P.]}, \bibinfo{author}{Thiran\xfnm[ J.P.]},
  \bibinfo{author}{Jonasson\xfnm[ L.]}, \bibinfo{author}{Vandergheynst\xfnm[
  P.]}, \bibinfo{author}{Clarke\xfnm[ S.]}, \bibinfo{author}{Maeder\xfnm[ P.]},
  \bibinfo{author}{Meuli\xfnm[ R.]}.
\newblock \bibinfo{title}{{DTI mapping of human brain connectivity: statistical
  fibre tracking and virtual dissection}}.
\newblock \emph{\bibinfo{journal}{NeuroImage}}
  \bibinfo{year}{2003};\bibinfo{volume}{19}(\bibinfo{number}{3}):\bibinfo{pages}{545--554}.
%Type = Book
\bibitem[{Kaufman and Rousseeuw(1990)}]{Kaufman1990}
\bibinfo{author}{Kaufman\xfnm[ L.]}, \bibinfo{author}{Rousseeuw\xfnm[ P.J.]}.
\newblock \bibinfo{title}{{Finding groups in data: An introduction to cluster
  analysis}}.
\newblock \bibinfo{publisher}{Wiley}; \bibinfo{year}{1990}.
%Type = Book
\bibitem[{Hastie et~al.(2009)Hastie, Tibshirani and Friedman}]{Hastie2009}
\bibinfo{author}{Hastie\xfnm[ T.]}, \bibinfo{author}{Tibshirani\xfnm[ R.]},
  \bibinfo{author}{Friedman\xfnm[ J.]}.
\newblock \bibinfo{title}{{The Elements of Statistical Learning}}.
\newblock \bibinfo{publisher}{Springer}; \bibinfo{year}{2009}.
%Type = Article
\bibitem[{Silva and Tenenbaum(2003)}]{Silva2003}
\bibinfo{author}{Silva\xfnm[ V.D.]}, \bibinfo{author}{Tenenbaum\xfnm[ J.B.]}.
\newblock \bibinfo{title}{{Global versus local methods in nonlinear
  dimensionality reduction}}.
\newblock \emph{\bibinfo{journal}{Advances in neural information processing
  systems}}
  \bibinfo{year}{2003};\bibinfo{volume}{15}:\bibinfo{pages}{705--712}.

\end{thebibliography}

\end{document}